\documentclass[runningheads]{ECCV/llncs}
\usepackage{graphicx}

\usepackage{tikz}
\usepackage{comment}
\usepackage{amsmath,amssymb} 
\usepackage{color}
\usepackage{epsfig}
\usepackage{graphicx}
\usepackage{multirow}
\usepackage{bm}
\usepackage{subcaption}
\usepackage{array}
\usepackage{xcolor}

\usepackage[accsupp]{axessibility}  

\newif\ifdraft
\drafttrue

\definecolor{orange}{rgb}{1,0.5,0}

\ifdraft
 \usepackage[normalem]{ulem}
 \newcommand{\RS}[1]{{\color{red}{\bf RS: #1}}}
 
 \newcommand{\PMN}[1]{{\color{orange}{\bf PMN: #1}}}

\else
 \newcommand{\sout}[1]{}
 \newcommand{\RS}[1]{{\color{red}{}}}
 
 \newcommand{\PMN}[1]{{\color{red}{}}}
 
\fi

\newcommand{\real}{\mathbb{R}}
\newcommand{\x}{\mathbf{x}}

\newcommand{\f}{\mathbf{f}}

\newcommand{\F}{\mathcal{F}}
\renewcommand{\a}{\mathbf{a}}

\newcommand{\MSCL}{{\bf{MSCL}}}

\newcommand{\SSD}{{\bf{SSD}}}
\newcommand{\DNtwo}{{\bf{DN2}}}
\newcommand{\MHRot}{{\bf{MHRot}}}
\newcommand{\DDV}{{\bf{DDV}}}
\newcommand{\IC}{{\bf{IC}}}
\newcommand{\HierAD}{{\bf{HierAD}}}
\newcommand{\Glow}{{\bf{Glow}}}
\newcommand{\MahaAD}{{\bf{MahaAD}}}
\newcommand{\CFlow}{{\bf{CFlow}}}

\newcommand{\uniclass}{\emph{uni-class}}

\newcommand{\uniano}{\emph{uni-ano}}
\newcommand{\unimed}{\emph{uni-med}}
\newcommand{\shiftlowres}{\emph{shift-low-res}}
\newcommand{\shifthighres}{\emph{shift-high-res}}

\begin{document}
\pagestyle{headings}
\mainmatter
\def\ECCVSubNumber{4951}  

\title{Data Invariants to Understand Unsupervised Out-of-Distribution Detection} 

\titlerunning{Data Invariants to Understand Unsupervised Out-of-Distribution Detection}
%
\author{Lars Doorenbos \and
Raphael Sznitman \and
Pablo Márquez-Neila}
\authorrunning{L. Doorenbos et al.}
%
\institute{
  University of Bern, Bern, Switzerland \\
  \texttt{\{lars.doorenbos,raphael.sznitman,pablo.marquez\}@unibe.ch}}
\maketitle

\begin{abstract}

Unsupervised out-of-distribution~(U-OOD) detection has recently attracted much attention due to its importance in mission-critical systems and broader applicability over its supervised counterpart. 
Despite this increased attention, U-OOD methods suffer from important shortcomings.
By performing a large-scale evaluation on different benchmarks and image modalities, we show in this work that most popular state-of-the-art methods are unable to consistently outperform a simple anomaly detector based on pre-trained features and the Mahalanobis distance~(MahaAD).
A key reason for the inconsistencies of these methods is the lack of a formal description of U-OOD.
Motivated by a simple thought experiment, we propose a characterization of U-OOD based on the invariants of the training dataset. 
We show how this characterization is unknowingly embodied in the top-scoring MahaAD method, thereby explaining its quality. Furthermore, our approach can be used to interpret predictions of U-OOD detectors and provides insights into good practices for evaluating future U-OOD methods.

\end{abstract}

\section{Introduction}

The use of deep learning~(DL) models for mission-critical systems, such as in autonomous driving or medicine, is one of the most active research areas in computer vision. Yet, despite impressive performances in recent methods, their ability to extrapolate beyond their training data remains limited. For trained and deployed models, this is particularly problematic when processing images that are corrupted or whose content differs from their expectation. Predictions for unexpected images are often incorrect with high confidence and cannot be identified as such~\cite{amodei2016concrete}. Ultimately, these silent failures deeply impact the reliability of machine learning systems in mission-critical applications and can have fatal consequences.

To mitigate these limitations, numerous \emph{out-of-distribution} (OOD) detection methods have emerged in the recent past. Closely related to anomaly detection~\cite{ruff2021unifying} and one-class learning~\cite{perera2021one}, OOD detection aims to spot samples at inference time that do not belong to the training distribution and should not be processed by subsequent machine learning models. At their core, OOD~detection methods learn scoring functions that measure the level of anomaly, or \emph{out-of-distributionness}, in test samples with respect to a training data distribution. 

\begin{figure*}[t]
    \centering
    \includegraphics[width=\linewidth]{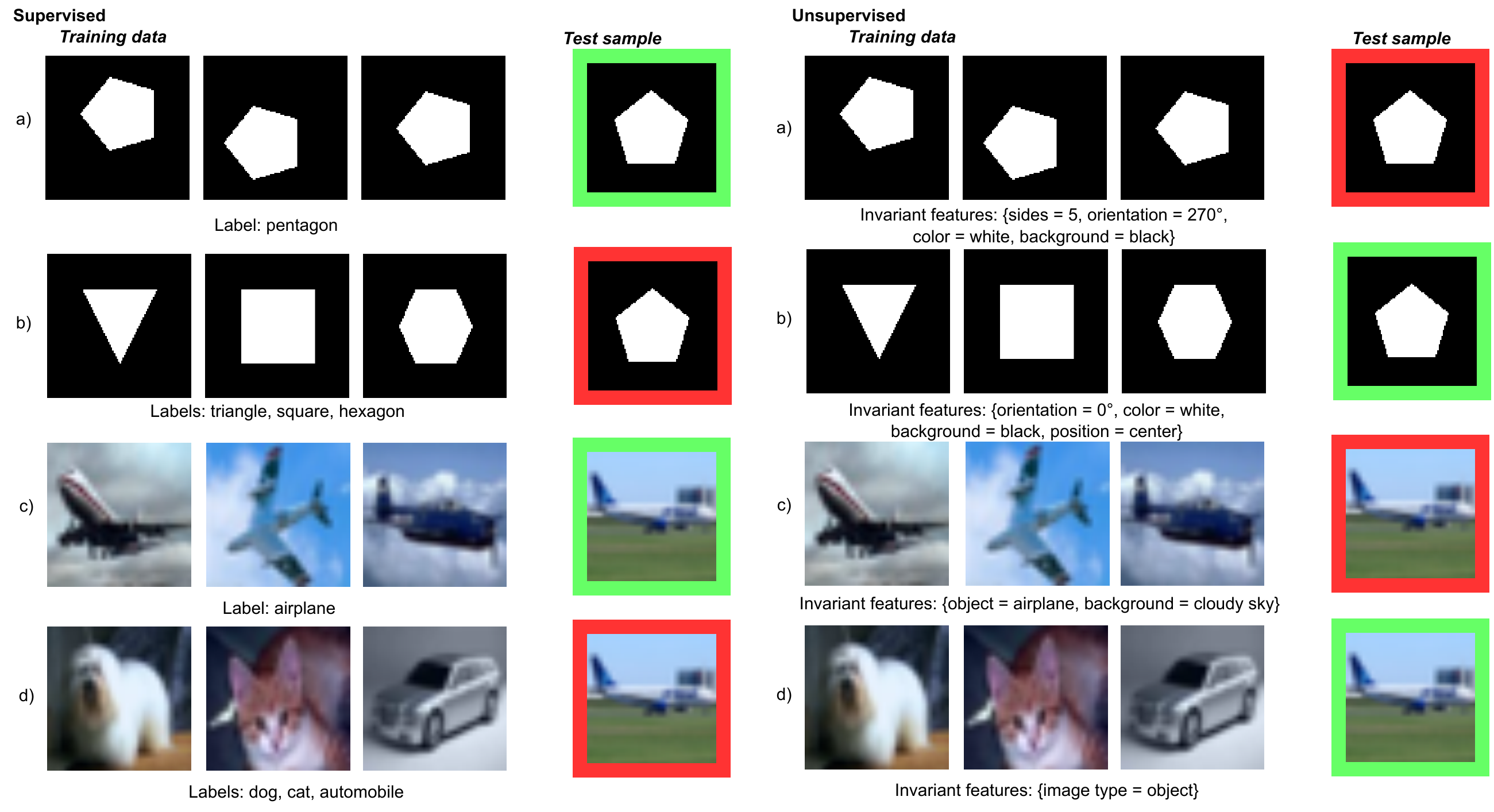}
    \caption{The difference between supervised and unsupervised OOD. For the unsupervised case, invariants in the training data rather than class labels define what should be considered as OOD: in (a)~a pentagon at a different angle leads to an OOD test sample, while (b)~shows variants in shapes in the training set such that a pentagon is in-distribution at test time. While the train and test data are the same in each row, the interpretation of OOD differs in the supervised and unsupervised cases. Green and red boxes denote in- and out-of-distribution samples, respectively.
    }
    \label{fig:geometric}
\end{figure*}

Broadly, OOD methods are categorized into supervised and unsupervised, as illustrated in Figure~\ref{fig:geometric}. Supervised OOD methods compute an OOD score by using the labels of the training dataset or by knowing the trained downstream network~\cite{hendrycks2016baseline,hsu2020generalized,kamoi2020mahalanobis}. Conversely, unsupervised OOD (U-OOD) methods are agnostic to the downstream task or data labels, and learn tractable representations of the training images to compute OOD scores \cite{bergman2020classification,choi2018waic,golan2018deep,ruff2018deep}, which makes them more general than supervised methods and applicable to a larger range of scenarios.

Considering its significance and generality, the recent emergence of U-OOD methods is unsurprising. Yet with many methods reporting state-of-the-art performance~\cite{havtorn2021hierarchical,hou2021divide,koner2021oodformer,morningstar2021density,reiss2021panda,salehi2021multiresolution,sehwag2021ssd,serra2019input,tack2020csi}, the overwhelming majority of these only validate their approach on one or two tasks. Given the broadness of U-OOD, these limited experimental validations have produced an inconsistent state-of-the-art, while simultaneously establishing an unclear sense of progress in the field. For instance,~\cite{hendrycks2019using} showed excellent results for one-class tasks using CIFAR10 and ImageNet, only to be contradicted 8 months later in~\cite{bergman2020deep} using different data. More alarmingly, this trend of inconsistencies is being perpetuated with evaluation protocols remaining unchanged~\cite{battikh2021latent,hou2021divide,mesarcik2021improving}.

For this reason, we first aim to explore and assess the performance and robustness of existing U-OOD detectors by establishing a wide and varied panel of experiments using different datasets and setups. Not only do we show that U-OOD state-of-the-art methods perform erratically when evaluated over a wide and varied range of datasets and tasks (\emph{i.e.} methods that perform extremely well on some datasets, frequently perform poorly on others), but that the relatively unnoticed MahaAD method~\cite{rippel2021modeling} consistently outperforms all considered methods by remarkable margins in addition to being extremely simple, stable, and easy to train. 

More fundamentally however, we hypothesize that despite the large number of recently proposed U-OOD methods, the main reason for this lack of overall consistency is that the fundamental concept of U-OOD remains vague and ill-defined. In fact, the vast majority of works fail to clearly define U-OOD, let alone provide an intuition to their approach's functioning. This subsequently leads to brittle methods and weak evaluation protocols. 

Intuitively, a test sample should be considered OOD if it \emph{looks different} from training samples. While this intuition seems straightforward, it is unclear how to characterize a training dataset or identify what makes a test sample similar or not to training samples. Yet, characterizing OOD is a fundamental necessity to not only produce reasonable U-OOD detectors, but also to properly evaluate and understand their behavior. Previous works have overlooked this important step and devised OOD detectors following more or less reasonable heuristics with limited formal justification. For example, using the observation that blurred images are assigned higher likelihoods compared to their original counterparts, SVD-RND~\cite{choi2019novelty} leveraged this property to characterize OOD by directly optimizing for it. Similarly,~\cite{ren2019likelihood,serra2019input} identified OOD samples by correcting for their input complexity and the number of background pixels. Other examples include Puzzle-AE~\cite{salehi2020puzzle}, which relied on solving puzzles of OOD images worse than their in-distribution counterparts, and MHRot~\cite{hendrycks2019using} assumed that geometric transformations of OOD samples will be predicted incorrectly. 

We also address here this apparent lack of a proper U-OOD definition by proposing a characterization based on identifying and leveraging image \emph{invariants} of the training set. Following this idea, we formulate the general problem of finding dataset invariants and show that, when constrained to a linear setting, this formulation reduces to the MahaAD method, which unknowingly embodies a dataset invariant characterization. Importantly, we show that the invariants found within a training set are more relevant for U-OOD detection than its variant counterparts.

In summary, the contributions of this paper include (1)~a thorough evaluation of numerous state-of-the-art U-OOD methods on different tasks and datasets, whereby highlighting that most methods perform erratically and inconsistently, (2)~a novel interpretation of U-OOD using training set invariants, which allows for an appropriate definition of U-OOD and (3)~a new U-OOD benchmark derived from our novel interpretation with invariants. A consequence of these contributions is that we shed light on why most recent methods do not perform well across datasets and, importantly, why the relatively unknown MahaAD method, that has been disregarded so far by most recent works in the field, is an excellent off-the-shelf U-OOD detector that should be included as a competitive baseline in future comparisons.


\section{Related works}
\label{sec:related}
Methods such as one-class support vector machines~\cite{scholkopf1999support}, isolation forest~\cite{liu2008isolation}, and local outlier factor~\cite{breunig2000lof} have traditionally been used for OOD detection in classical machine learning. However these methods suffer greatly when applied to high-dimensional spaces (\emph{i.e.} images). Unsurprisingly, DL based methods have come to replace these more recently. Summarized here are some of the most relevant works on OOD~detection using DL, while comprehensive surveys can be found in~\cite{chalapathy2019deep,yang2021generalized}.

Supervised OOD detection approaches require either an explicitly trained classifier or a labelled dataset to work. One line of works uses a classifier's maximum softmax probability output as the OOD score~\cite{hendrycks2016baseline,hsu2020generalized,liang2017enhancing}. Another, more closely related to U-OOD, exploits deep features of the task-specific trained classifiers~\cite{kamoi2020mahalanobis,lee2018simple,sastry2020detecting}. However, as all these methods exploit relations between network predictions and the path taken to arrive at those predictions in some way, they are simply incompatible with the U-OOD setting.

On the other hand, U-OOD detection methods rely only on a set of {\it in}-distribution images to learn the characteristics of the  in-distribution data. That is, they do not assume, or have access to, a trained downstream deep network or labeled dataset. Broadly, two families of methods are found in the literature. The first are generative models while the second are based on representation learning. 

{\bf Generative models:} These learn the distribution of images in high-dimensional spaces. However, most generative models are known to perform poorly in OOD~detection~\cite{choi2018waic,nalisnick2018deep}, and many augmentations and improvements have been proposed to increase their performances.~\cite{serra2019input} showed that the likelihoods obtained by models such as Glow~\cite{kingma2018glow} or PixelCNN++~\cite{salimans2017pixelcnn++} are heavily influenced by the input complexity, and propose a likelihood ratio to correct for this. Interestingly, the work in \cite{ren2019likelihood} showed that background pixels dominate test sample likelihood scores, and attempt to correct for these by using the likelihood of a second model that tries to capture the population level background information. Similarly, Schirrmeister \emph{et al.}~\cite{schirrmeister2020understanding} use the likelihood ratio with respect to a second model trained on a general, large scale dataset.

{\bf Representation learning:} 
Instead of working in the image space, most U-OOD methods aim to learn a low-dimensional image embedding. Here, many works have opted for self-supervised learning strategies to simulate classification problems and train DL models to representative image features. One popular approach is predicting geometric transformations, such as image rotations, translations, scales, flips, or patch re-arrangements~\cite{bergman2020classification,golan2018deep,hendrycks2019using,wang2019effective}. Other self-supervised approaches rely on auto-encoders and optionally perturb the input in some way to create more robust feature descriptions. Example perturbations include adding noise~\cite{sakurada2014anomaly} or shuffling patches~\cite{salehi2020puzzle}. Further extensions propose to fit an auto-regressive model to the latent space~\cite{abati2019latent} or to add a memory module~\cite{gong2019memorizing}. Most recently, approaches based on contrastive learning have been advantageous~\cite{reiss2021mean,sehwag2021ssd,tack2020csi}.

However, various papers showed that learning features on the target domain is not necessary to reach high performance \cite{bergman2020deep,ouardini2019towards,rippel2021modeling,xiao2021we}. Bergman \emph{et al.}~\cite{bergman2020deep} find that scoring samples by the distance to their k-nearest neighbours in the space of pre-trained ImageNet features outperformed all previous self-supervised methods. Xiao \emph{et al.}~\cite{xiao2021we} showed that exploiting features obtained from self-supervised ---rather than supervised--- training on ImageNet can lead to to high performance. Finally, Rippel \emph{et al.}~\cite{rippel2021modeling} combined Mahalanobis distances in the space of ImageNet features for state-of-the-art results on the MVTec dataset. 
\section{Invariants for Unsupervised OOD}
\label{sec:method}

In the supervised setting, similar to the problem of zero-shot learning, a sample is considered OOD if it cannot be assigned to one of the training set classes. In the unsupervised setting, however, defining OOD is more challenging as we do not know {\it a-priori} what and  if any classes are present at all. As done in anomaly detection~\cite{ruff2021unifying}, one potential approach to define U-OOD could be to measure if a sample lies in a low-density region of the training data. But doing so would be inappropriate because whether few or many image examples of a specific class appear in a training set may only be a reflection of their natural prevalence, rather than being a real OOD sample. For instance, if one had a training set of dogs, the Norwegian Lundehund (\emph{i.e.} a rare dog breed) would most likely appear in low-density regions of the training distribution, in contrast to German Shepherds (\emph{i.e.} very common bread). Yet both should still be considered dogs. Instead, we propose to use \emph{invariants} as a way to characterize U-OOD. Specifically, our idea is to first determine image invariants in the training set, and then detect OOD test samples by identifying if they keep the invariants of the training set.

To illustrate this, consider the toy examples in Fig.~\ref{fig:geometric}, where four different combinations of training sets and test examples are given. Recall that for the unsupervised case, no labels in the training data are available thus losing context as to what is or is not semantically OOD. However, the necessity to leverage context to disentangle relevant and irrelevant aspects of images remains key for U-OOD detection, since it is too broad to be meaningful without it (as stated in~\cite{ahmed2020detecting}). Hence, we assume that this necessary context is provided by a set of general features that we have at our disposal, that can describe the input images~$\x$. For instance, these features could be $\f(\x)=\{\textrm{sides}(\x), \textrm{orientation}(\x), \textrm{color}(\x), ...\}$, or features coming from a network pre-trained on a general dataset. Given this, we want to summarize a training set by the {\it union} of features that are invariant over the entire training set. For example, Fig.~\ref{fig:geometric}(a) would use the combination of invariant features $\{\textrm{sides}=\textrm{5}, \textrm{orientation} = 270^\circ, \textrm{color}=\textrm{white}, \textrm{background}=\textrm{black}\}$, and similarly $\{\textrm{orientation}=0^\circ, \textrm{color}=\textrm{white}, \textrm{background}=\textrm{black}, \textrm{position}=\textrm{center}\}$ for Fig.~\ref{fig:geometric}(b). At inference time then, a test sample described by this union of invariant features would be OOD if these features are no longer invariant with respect to the training set. In this sense, variant features from a dataset are in fact irrelevant for U-OOD detection, which stands in contrast to many previous methods that focused on learning a representation of the training distribution (\emph{e.g.},~\cite{choi2019novelty,marquez2019image,zong2018deep}). 

In the remainder of this section, we begin by formalizing the above-mentioned idea and propose an approach to identifying these invariants for the general case. We then show how this is related to the MahaAD method~\cite{schirrmeister2020understanding}. In the experimental section, we demonstrate how MahaAD performs in comparison to recent methods and how it behaves in light of image invariants.  

\subsection{Formalization}

Given a training set~$\{\x_i\}_{i=1}^N$, with corresponding feature vectors,~$\f(\x_i) \equiv \f_i \in \F$, we define an invariant as a non-constant function~$g:\F\to\real,$ such that $g(\f_i)=0,\ \forall i$. That is, $g$~is an invariant if it computes a constant value (\emph{i.e.},~$g(\f_i)=0$) for the elements of the training set, but in general may not compute the same constant value for other elements (\emph{e.g.},~elements of a test set). Our goal then is to find a set of invariants,~$G=\{g_1, \ldots, g_K\}$, over the set of training feature vectors. While doing so in one global optimization is challenging, we propose to solve this  by solving a sequence of $K$~problems, one per invariant,
\begin{align}
    \label{eq:solve_g}
    g_k(\f_i) &= 0 \quad \forall i, \\
    \nonumber
    \|\nabla g_k(\f_i)\|_2 &\neq 0 \quad  \forall i,\\
    \nonumber
    \nabla g_k(\f_i)\cdot \nabla g_j(\f_i) &= 0 \quad  \forall i, j<k,
\end{align}
where the first equality makes~$g_k$ zero for all training samples, the second equality prevents~$g_k$ from becoming a projection (\emph{i.e.},~effectively making it non-constant) and the third equality requires that new invariants are different from all previously found invariants by making their gradients mutually orthogonal. After finding~$G$, a test feature vector~$\f$ will be considered OOD if $g_k(\f) \neq 0$ for any invariant~$k$.

As noisy real-world data rarely lies in an exact manifold, solving Eq.~\eqref{eq:solve_g} is unfeasible in practice even for a small number of invariants~$K$. Instead, we relax Eq.~\eqref{eq:solve_g} and express it as a minimization problem to find a set of soft invariants,
\begin{align}
    \label{eq:min_g}
    \min_{g_k} & \dfrac{1}{N}\sum_i g_k(\f_i)^2, \\
    \nonumber
    \textrm{s.t.}   &\quad\|\nabla g_k(\f_i)\|_2 = 1 \quad  \forall i, \\
    \nonumber
                    &\quad \nabla g_k(\f_i)\cdot \nabla g_j(\f_i) = 0 \quad \forall i, j<k,
\end{align}
where we constrain the magnitude of the gradient to~$1$ to prevent~$g_k$ from arbitrarily compressing its output and minimizing the loss artificially. 

Once $G=\{g_1, \ldots, g_K\}$ is established, any test vector~$\f$ can be scored by computing the ratios between the test error and the average training error,
\begin{equation}
    \label{eq:score_g}
    s^2(\f) = \sum_k\dfrac{g_k(\f)^2}{e_k},
\end{equation}
where $e_k$ is the training MSE~of the soft invariant~$g_k$,
\begin{equation}
    e_k = \dfrac{1}{N}\sum_i g_k(\f_i)^2.
\end{equation}
Intuitively, tight invariants with low~$e_k$ values will have a high influence in the final score, while weak invariants with large~$e_k$ values will essentially be ignored. Given that the contribution of weak invariants is negligible in~$s^2$, we can circumvent the problem of setting an optimal number of invariants~$K$ and safely set~$K$ to the dimensionality of the feature space.

We can further simplify the optimization problem of Eq.~\eqref{eq:min_g} by constraining the invariants to the family of affine functions~$g_k(\f) = \a_k^T\f + b_k$ with unitary~$\a_k$. Under these conditions, Eq.~\eqref{eq:min_g} reduces to a PCA problem. Its solution sets~$\a_k$ to the k-th smallest principal component and the squared error~$e_k$ is set to its corresponding eigenvalue. Moreover, the score function Eq.~\eqref{eq:score_g} can be re-written as the square of the Mahalanobis distance using the mean and the covariance of the training feature vectors. Ultimately, computing Mahalanobis distances properly weighs and exploits the linear invariants in the training dataset, which, in turn, suggests that the Mahalanobis distance could lead to good OOD~detectors despite its simplicity.

Given that the invariants are computed, in practice, from a collection of feature vectors describing the training set, the performance of an invariant-based U-OOD detection method is contingent on the chosen pre-trained feature extractor. We experimentally found that this is not an important limitation and that general ImageNet-based features lead to descriptive invariants for U-OOD detection even when applied on image modalities that are very different from ImageNet, such as medical images.

\subsection{The Mahalanobis anomaly detector}

\begin{figure}[t]
    \centering
    \includegraphics[width=0.95\linewidth]{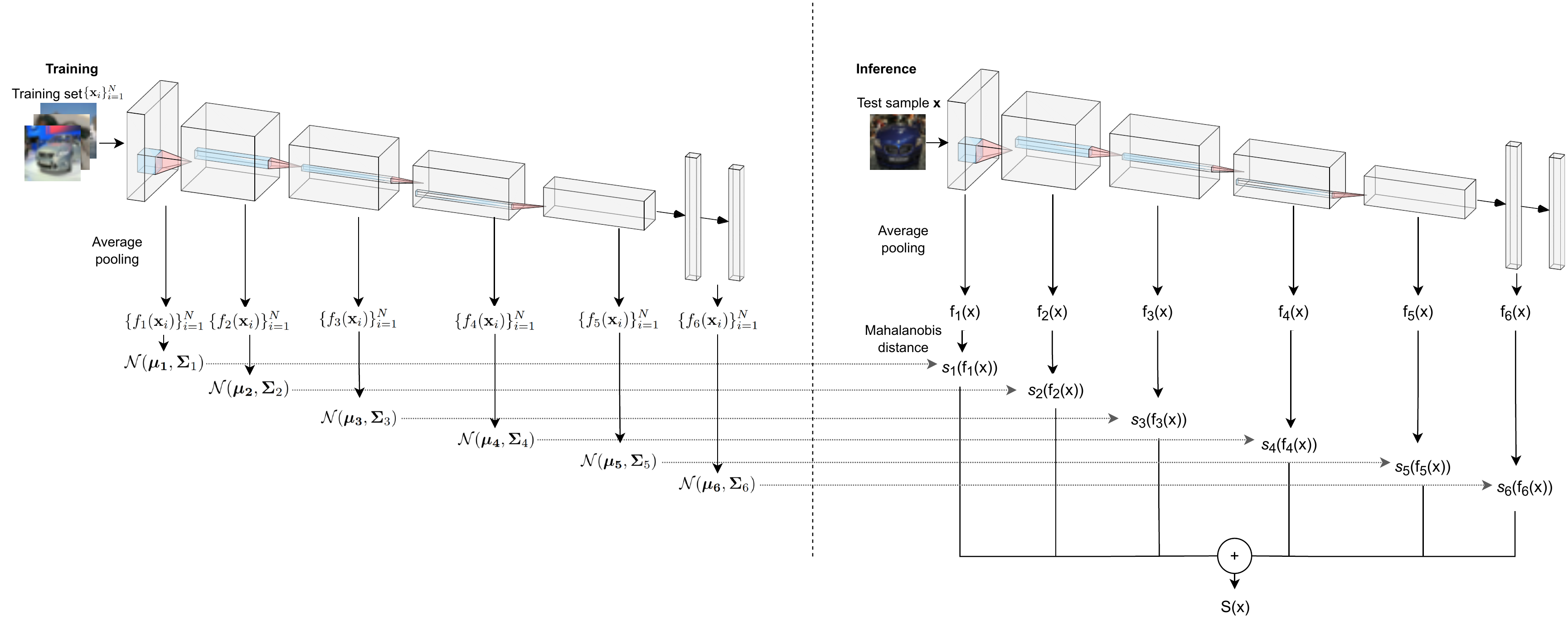}
    \caption{Training and inference stages of the MahaAD method.}
    \label{fig:mahaad}
\end{figure}

Given the above, we briefly revisit the the Mahalanobis anomaly detector (MahaAD) from Rippel \emph{et al.}~\cite{rippel2021modeling} as it embodies the invariant feature learning we propose. Fig.~\ref{fig:mahaad} illustrates the approach. 

MahaAD uses the spatial pooling of the feature maps of a pre-trained CNN to define feature descriptors~$\f$. Instead of choosing a specific CNN layer for~$\f$, MahaAD works in a multi-layered manner describing each input image~$\x$ with a collection of feature vectors~$\{\f_\ell(\x)\}_{\ell=1}^L$ computed at $L$~different layers.

At training time, MahaAD computes the mean and the covariance of the descriptor vectors of the images in the training dataset~$\{\x_i\}_{i=1}^N$. Specifically, for each layer~$\ell$, the mean is computed as,
\begin{equation}
    \bm{\mu}_{\ell} = \frac{1}{N}\sum_{i=1}^N \f_{\ell}(\x_i),
\end{equation}
while the corresponding covariance matrix is,
\begin{equation}
    \bm{\Sigma}_{\ell} = \frac{1}{N}\sum_{i=1}^N (\f_{\ell}(\x_i) - \bm{\mu}_l)(\f_{\ell}(\x_i) - \bm{\mu}_l)^\intercal.
\end{equation}
To avoid singular covariance matrices in high-dimensional or low-data regimes, shrinkage is applied using the standard hyperparameter-free method of~\cite{ledoit2004well}, although we empirically found that the shrinkage has limited impact on the overall performance of MahaAD. 
By using multi-layer feature vectors, MahaAD is able to find linear invariants at different image scales.

Importantly, the CNN from which the features are computed is not trained or tuned to the training set whereby making this training phase simple and extremely fast. In practice, it makes the approach more stable and robust across a larger number of datasets. This differs from most recent U-OOD methods that opt to fine-tune their DL models to the training set~\cite{reiss2021panda,sehwag2021ssd,tack2020csi}. 

At test time, MahaAD computes the layer-wise Mahalanobis distances between the descriptor vectors of the test image~$\x$ and the means~$\{\bm{\mu}_\ell\}_\ell$,
\begin{equation}
 s_\ell(\f) 
 = \sqrt{(\f - \bm{\mu}_{\ell})^\intercal \bm{\Sigma}_{\ell}^{-1} (\f - \bm{\mu}_{\ell})},
\end{equation}
which is equivalent to the square root of Eq.~\eqref{eq:score_g}. The final OOD score is the sum of the scores over all layers, 
\begin{equation}
    S(\x) = \sum_{\ell=1}^L s_{\ell}(\f_\ell(\x)).
\end{equation}
\section{Experiments}
\label{sec:experiments}

To explore the current state of U-OOD detection, we design a benchmark comparing the performance of several state-of-the-art U-OOD methods over a broad collection of 73~experiments that involve different image modalities, sizes, perturbations, and different criteria for the in- and out-distributions. These experiments aim to identify in what scenarios different methods may be effective and which may not be. Our benchmark is organized in five tasks (see Fig.~\ref{fig:datasets}):

\begin{description}
    \item[Unimodal CIFAR (\uniclass).] Similar to most works~\cite{hendrycks2019using,hou2021divide,koner2021oodformer,reiss2021mean,salehi2021multiresolution,sehwag2021ssd,tack2020csi}, we perform 10 experiments using the CIFAR10 dataset, where each experiment takes one of the 10 classes as in-distribution and uses the remaining 9 as OOD. We also use CIFAR100 for 20~experiments, where each of the 20~semantic superclasses of CIFAR100 are used as in-distribution and treat all remaining 19~superclasses as~OOD~\cite{bergman2020deep,golan2018deep,reiss2021mean,tack2020csi}.
    \item[Unimodal anomaly (\uniano).] We use the MVTec dataset~\cite{Bergmann_2019_CVPR} which contains 15~classes of images of both normal and defect objects. As in~\cite{defard2021padim,li2021cutpaste,reiss2021mean,rippel2021modeling,sohn2020learning}, we perform one experiment per class, where the defect-free images are used for the in-distribution and defect test images are considered~OOD samples.
\begin{figure}[!b]
    \centering
    \includegraphics[width=\linewidth]{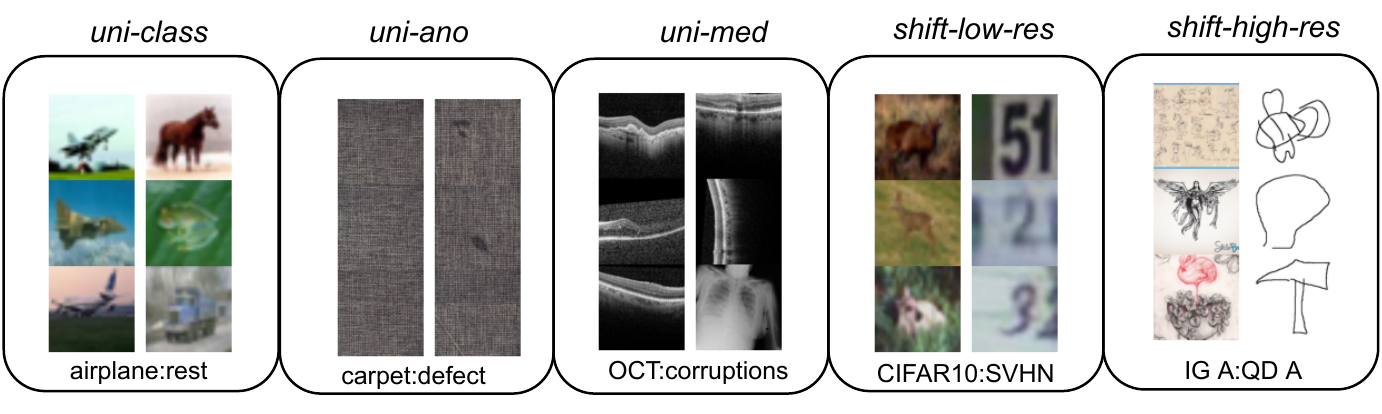}
    \caption{Example in-distribution and OOD images for each task in our proposed benchmark. IG stands for infograph, QD for quickdraw.}
    \label{fig:datasets}
\end{figure}
    \item[Unimodal anomaly medical (\unimed).] We perform 7~experiments with different medical image modalities. The first 2 experiments use optical coherence tomography~(OCT) scans and chest X-rays as training in-distributions and corrupted images as OOD samples. The 3\textsuperscript{rd} experiment trains the models with healthy chest X-rays and uses pathological chest X-rays as~OOD. In the remaining 4 experiments, healthy retinal fundus photographs are used for the in-distribution and pathological fundus photographs of four increasing severity levels are used for the OOD images~\cite{bozorgtabar2020salad,marquez2019image,ouardini2019towards,tang2019abnormal}.
    \item[Low-resolution domain shift (\shiftlowres).] 1~experiment using CIFAR10 as the in-distribution and SVHN as~OOD~\cite{mohseni2021multi,morningstar2021density,sehwag2021ssd,serra2019input,tack2020csi}. In contrast to previous works, we do not consider CIFAR100 as~OOD.
    \item[High-resolution domain shift (\shifthighres).] An extended version of the experiments on the dataset DomainNet presented by Hsu et al.~\cite{hsu2020generalized}. We run 20~experiments separated into two groups: 10~experiments with \texttt{Real-A} as the in-distribution and 10~experiments with \texttt{Infograph-A} as the in-distribution. We avoid using \texttt{Real-B} and \texttt{Infograph-B} as OOD in the first and the second group of experiments respectively.
\end{description}
We refer to a specific experiment by the notation in-dataset:out-dataset. 


\textbf{Evaluated methods}: We evaluate a selection of top-performing U-OOD~detection methods, which we selected if they were among the top-performing methods in at least one of the above tasks. These are: \MSCL~\cite{reiss2021mean},~\DNtwo~\cite{bergman2020deep}, \SSD~\cite{sehwag2021ssd}, \MHRot~\cite{hendrycks2019using}, \textbf{DDV}~\cite{marquez2019image}, \Glow~\cite{kingma2018glow}, \IC~\cite{serra2019input}, \HierAD \cite{schirrmeister2020understanding} and \CFlow~\cite{gudovskiy2022cflow}.

All methods are used with their default hyperparameters as given in their official implementations, with (where applicable) the same backbone architecture. More details can be found in the supplementary material. No hyperparameter search was performed, given that no validation metrics exist. 
Specifically, \Glow, \HierAD~and \IC~models are based on the same Glow network. All other methods use a ResNet-101. All methods resize the input images to $224 \times 224$, with the exception of the Glow-based and \SSD~models, where we found that resizing images to $32 \times 32$ worked better.

Additionally, we show results when using \MahaAD~with an EfficientNet-b4. As \MahaAD~requires no neural network training and thus, unlike all other methods except \DNtwo, the additional computational cost for this is minimal. Note that using a pre-trained CNN to extract image features is not a major limitation in practice, as all standard deep learning libraries offer tools to load and use such models in very few steps. 

\subsection{Results}

The evaluated methods were compared in terms of performance, training times and training complexity. We detail the results of our experiments below and provide a deeper breakdown of the results, including additional methods, in the supplementary material.

\begin{table*}
\caption{Performance summary in AUC over three runs on our U-OOD benchmark. We report performances for each task, as well as the mean over tasks and over experiments. No standard deviation is reported for \MahaAD~and~\DNtwo~as they are deterministic. ($^*$) Taken from original publication; ($^+$) taken from~\cite{reiss2021mean}; ($^-$) taken from~\cite{tack2020csi}; ($^\dagger$) taken from~\cite{schirrmeister2020understanding} }  
\label{tab:best}
\begin{center}
\resizebox{\textwidth}{!}{
\begin{tabular}{>{\bfseries}lc|ccccc|cc}

\multirow{2}{*}{Method}    & \multirow{2}{*}{\textbf{Architecture}} & \textbf{\uniclass}& \textbf{\uniano}      & \textbf{\unimed}        & \textbf{\shiftlowres}   & \textbf{\shifthighres}              & \textbf{Task} &\textbf{Experiment}   \\
&&&&&&&\textbf{Mean}&\textbf{Mean} \\
\hline
Glow & K = 32, L = 3      & 53.8\tiny{$\pm$0.1} &82.0\tiny{$\pm$2.5}  &55.8\tiny{$\pm$0.8}    &8.8 $^\dagger$                    & 34.5\tiny{$\pm$0.1}  & 47.0 & 53.9  \\
IC &K = 32, L = 3 &55.7\tiny{$\pm$0.1}  & 73.6\tiny{$\pm$2.6} &65.1\tiny{$\pm$0.5}    &95.0$^\dagger$                    & 65.8\tiny{$\pm$0.1} &71.0 & 63.6   \\
HierAD & K = 32, L = 3& 63.0\tiny{$\pm$0.4} &81.6\tiny{$\pm$2.1}  &72.5\tiny{$\pm$0.6}    &93.9$^*$                 & 75.0\tiny{$\pm$0.3}  &77.2 & 71.4  \\
MHRot & ResNet-101 & 83.4$^+$ & 70.8\tiny{$\pm$1.0} &69.0\tiny{$\pm$0.7}   & \underline{97.8$^-$} & 73.3\tiny{$\pm$0.9} &78.9 & 76.9 \\
DDV   & ResNet-101   & 65.8\tiny{$\pm$1.4}  &65.5\tiny{$\pm$0.2}&60.3\tiny{$\pm$3.2}& 47.9\tiny{$\pm$6.6}& 63.9\tiny{$\pm$4.9} & 60.7& 64.5\\
MSCL & ResNet-101 & \textbf{96.3\tiny{$\pm$0.0}} & 86.4\tiny{$\pm$0.0} & \underline{75.2\tiny{$\pm$0.1}} & 88.3\tiny{$\pm$0.0} &74.4\tiny{$\pm$0.0} & \underline{84.1} & \underline{86.1}\\
CFlow & ResNet-101 &75.0\tiny{$\pm$0.0}&\textbf{95.7\tiny{$\pm$0.1}} &68.8\tiny{$\pm$0.3} &6.6\tiny{$\pm$0.2} & 61.8\tiny{$\pm$0.3} & 61.6 & 74.1\\
DN2 & ResNet-101 &91.2&86.2&\textbf{76.7}&57.4&\underline{76.0}&  77.5 & 84.1\\
SSD & ResNet-101 &83.6\tiny{$\pm$0.3}& 65.8\tiny{$\pm$3.0} &64.6\tiny{$\pm$0.6}   & \textbf{99.6$^*$}       & 60.4\tiny{$\pm$0.9}  &74.8 & 72.0 \\
\hline
MahaAD & ResNet-101  & \underline{92.4} & \underline{91.3} & 75.7 &94.3& \textbf{78.6}& \textbf{86.5} & \textbf{86.8} \\
\textcolor{gray}{MahaAD} & \textcolor{gray}{EfficientNet-b4}     & \textcolor{gray}{95.1}     &  \textcolor{gray}{94.4} & \textcolor{gray}{76.8} & \textcolor{gray}{96.2}& \textcolor{gray}{83.8} & \textcolor{gray}{89.3} & \textcolor{gray}{90.1}\\
\end{tabular}}
\end{center}
\end{table*}

\begin{figure}[b!]
    \centering
    \setlength\tabcolsep{16pt}
    \begin{tabular}{cc}
        \includegraphics[width=0.33\linewidth]{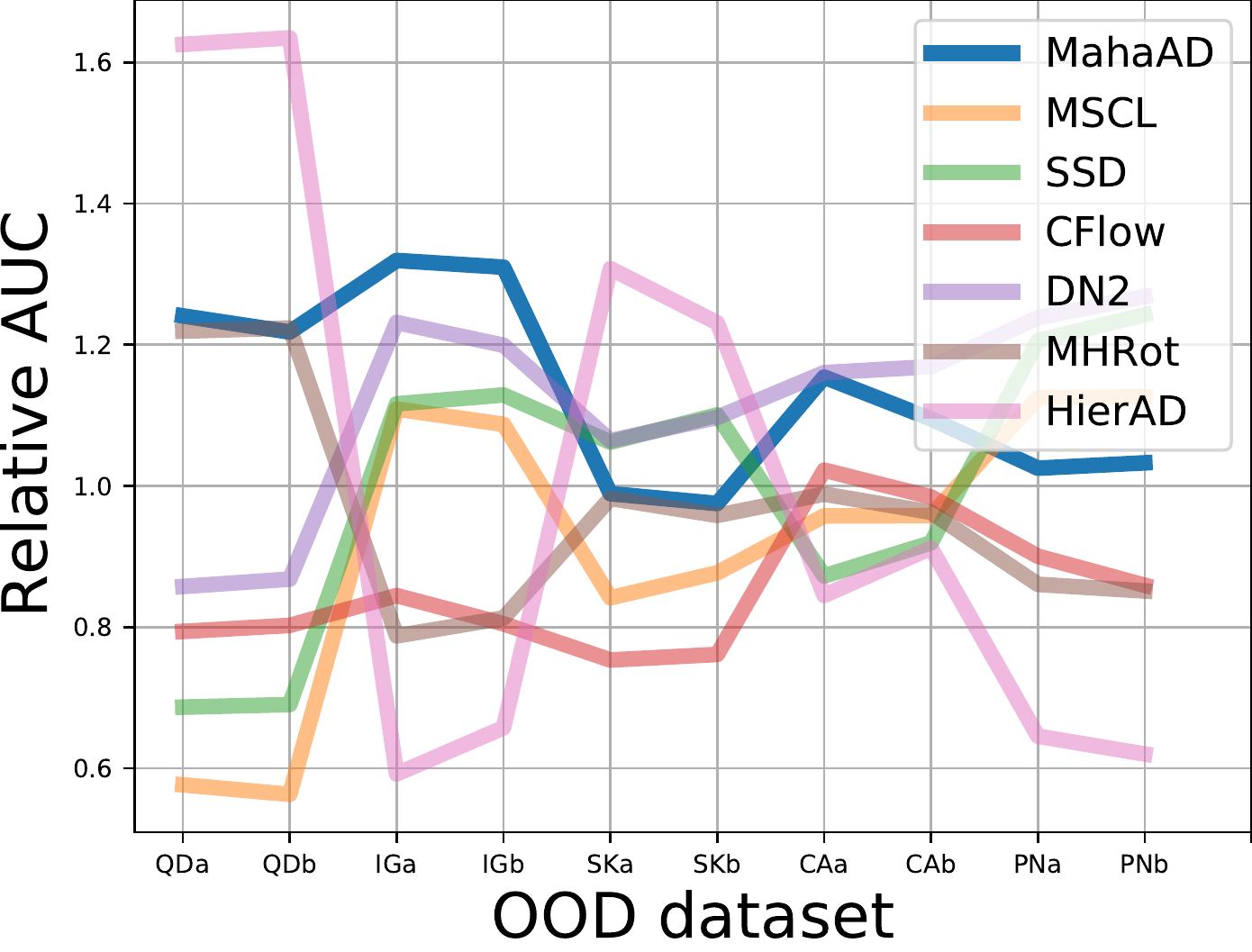} &
        \includegraphics[width=0.33\linewidth]{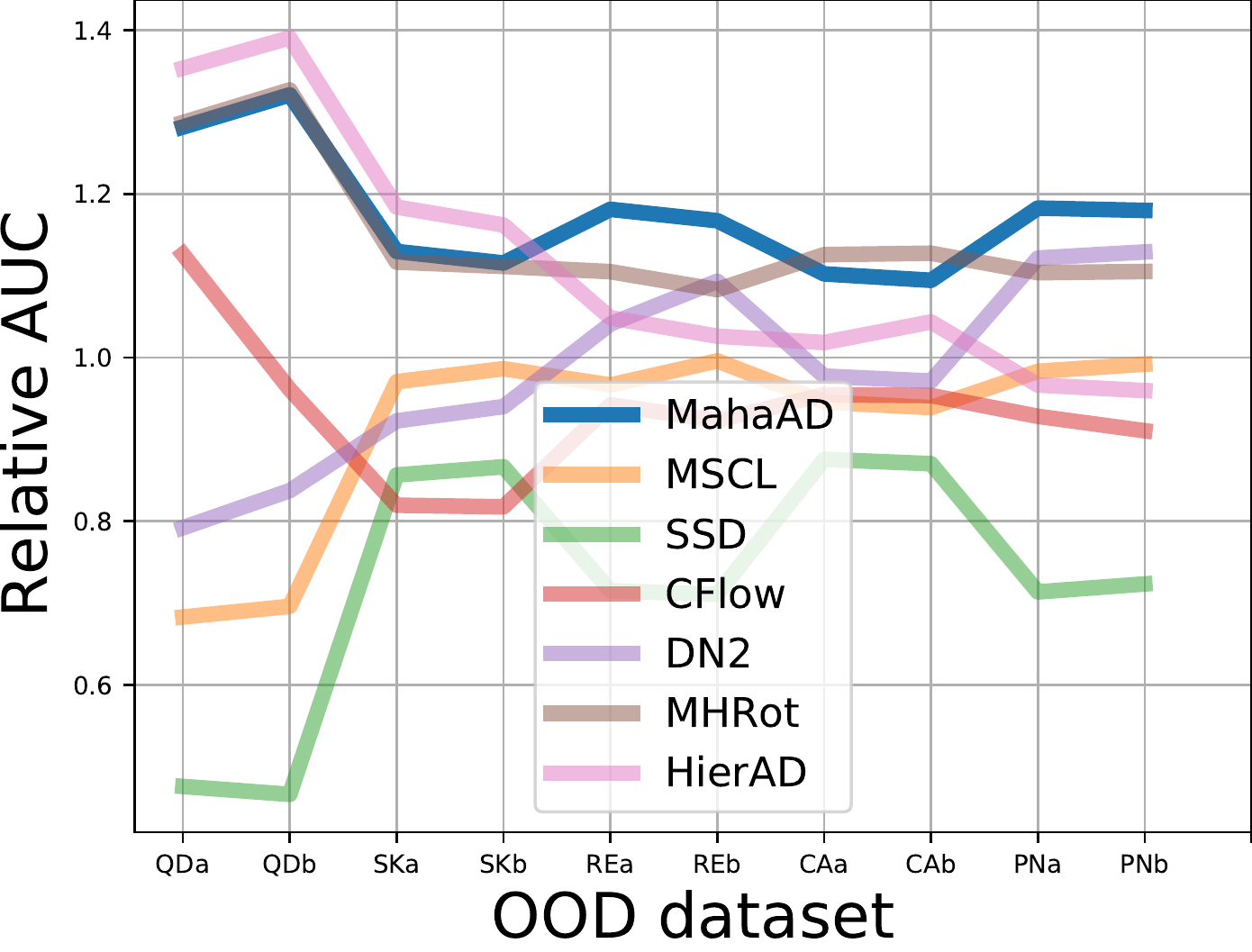} \\
        (a) & (b) \\
    \end{tabular}
    \caption{Relative performance (AUC divided by mean~AUC on that task) for seven methods on the \shifthighres~tasks. X-axis indicates the out distribution. (a)~\texttt{Real-A} as in-distribution. (b)~\texttt{Infograph-A} as in-distribution.}
    \label{fig:dn}
\end{figure}

\textbf{Performance.} Most methods were inconsistent across different tasks (see Table~\ref{tab:best}). \MSCL, which performed very well in \uniclass, is challenged in \uniano~and in \shiftlowres. Conversely, \CFlow's performance is high for
\uniano, but heavily drops in \uniclass~and especially \shiftlowres. \SSD~had the best results on \shiftlowres~but struggled with tasks involving high-dimensional images, and \DNtwo~scored very well on average except on \shiftlowres.
On the other hand, \MahaAD~performed very well and with high stability across tasks. Specifically, it~performed among the top three methods in all tasks but in the low resolution domain shift task, for which it still beats \MSCL, \DNtwo~and~\DDV~by large margins. Furthermore, \MahaAD~was the best performing method on average, beating the second-best method, \MSCL, by more than 2~percent points across tasks. 
These performance instabilities were not only observed across the different tasks reported in Table~\ref{tab:best}, but also within the tasks with fixed in-distribution across different OOD~datasets. For example, for the \shifthighres~task, performance of most methods fluctuated depending on the chosen OOD~dataset~(see Fig.~\ref{fig:dn}). In contrast, \MahaAD~again is the only method that stands out in terms of stability, as it performs well regardless of the in and the out datasets selected.

\begin{figure}[b!]
    \centering
    \setlength\tabcolsep{16pt}
    \begin{tabular}{cc}
        \includegraphics[width=0.33\linewidth]{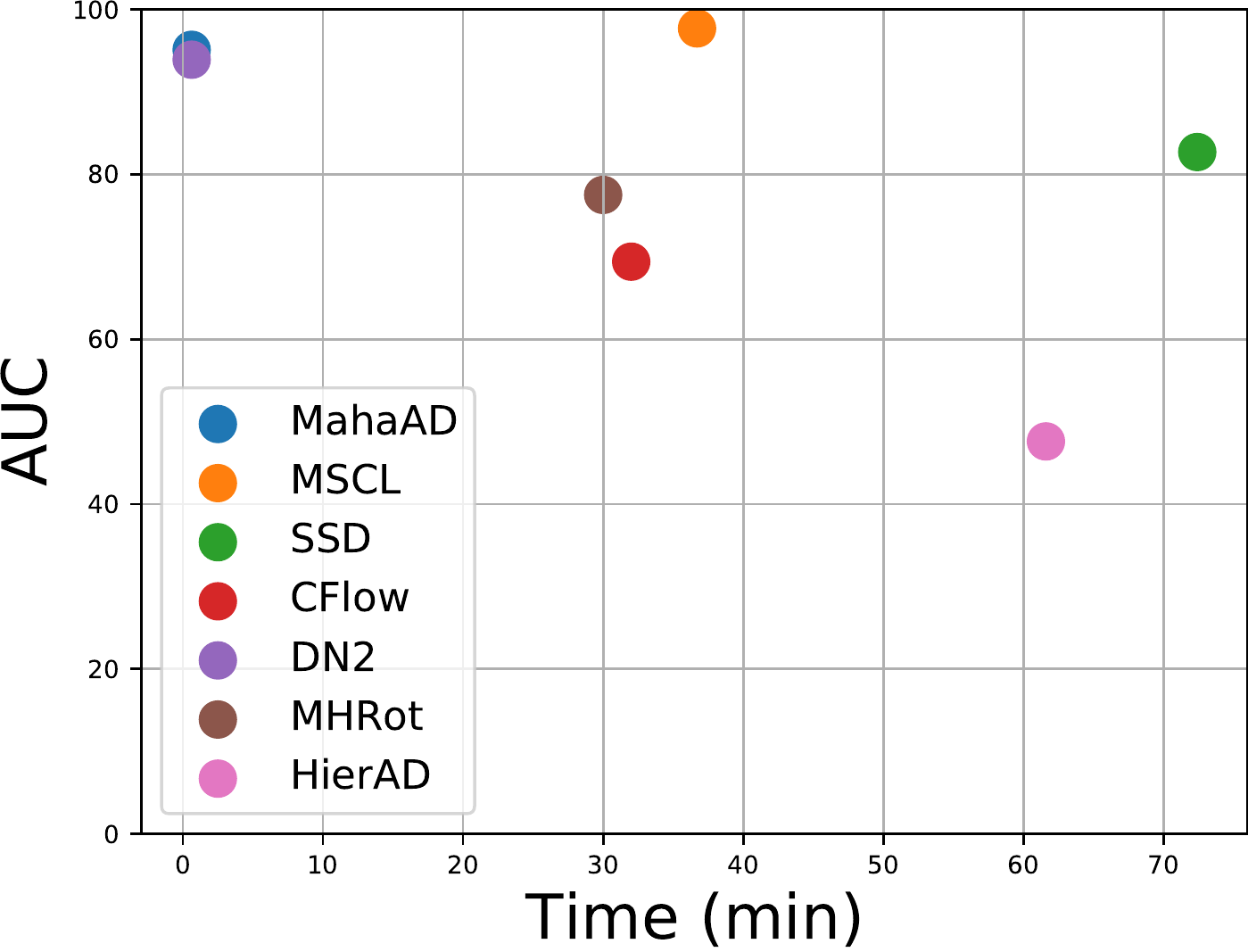} &
        \includegraphics[width=0.33\linewidth]{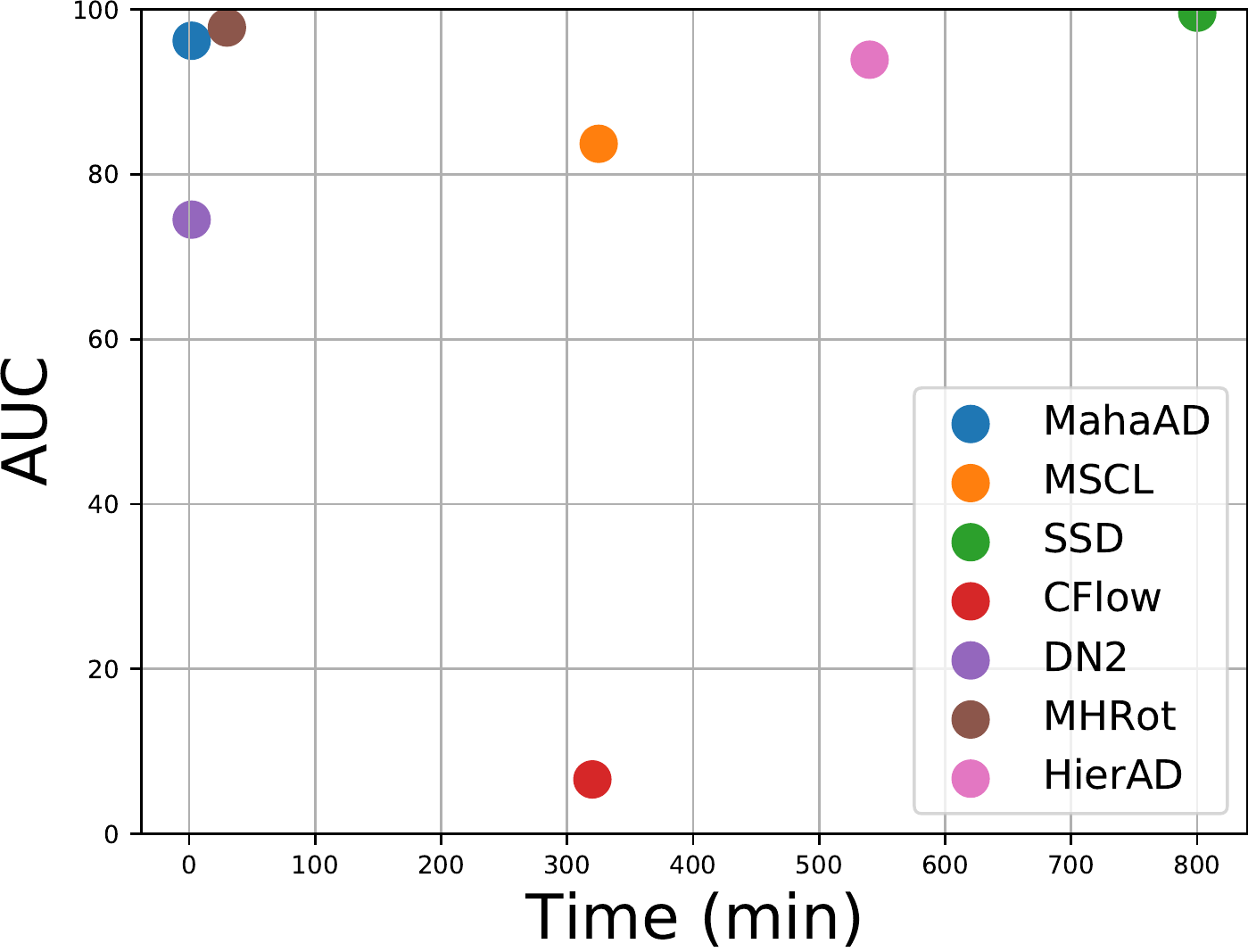} \\
        (a) & (b) \\
    \end{tabular}
    \caption{Training times and performances for different methods on (a) \uniclass's \texttt{airplane:rest} and (b) \shiftlowres's \texttt{CIFAR10:SVHN}.}
    \label{fig:times}
\end{figure}

\textbf{Training times.} \MahaAD~was faster to train than its counterparts~(see Fig.~\ref{fig:times}). For example, in the CIFAR10:SVHN experiment (task \shiftlowres), using two GeForce RTX 3090s, \MahaAD~was the fastest to train, taking roughly 90 seconds to process the entire CIFAR10~dataset. Other methods with similar performances were orders of magnitude slower: \MSCL~took more than half an hour for airplane:rest and \SSD~took more than 12~hours for~CIFAR10:SVHN. In addition, no method performed consistently better than \MahaAD~on either of these two experiments. This behavior was also observed for the other tasks.

\textbf{Training complexity.}
Furthermore, \MahaAD~was simpler to train, with fewer hyperparameters and more predictable behavior. Predicting the convergence of methods such as \MSCL, \CFlow~and \DDV~was challenging as there is no apparent correlation between the training loss and OOD~performance, as also reported in~\cite{reiss2021mean}. It is thus unclear when to stop training before the performance starts degrading. While this lack of obvious stopping criterion is problematic for many methods~\cite{perera2019ocgan,reiss2021panda,reiss2021mean,ruff2018deep}, \MahaAD~is convenient as it avoids this necessity altogether.

\subsection{Importance of data invariants}

We report here additional results that support the importance of data invariants, both for the quality of U-OOD detection and as a tool to analyse U-OOD predictions and evaluation datasets.

\begin{figure}[t]
  \centering
  \setlength\tabcolsep{16pt}
  \begin{tabular}{cc}
    \includegraphics[width=0.33\linewidth]{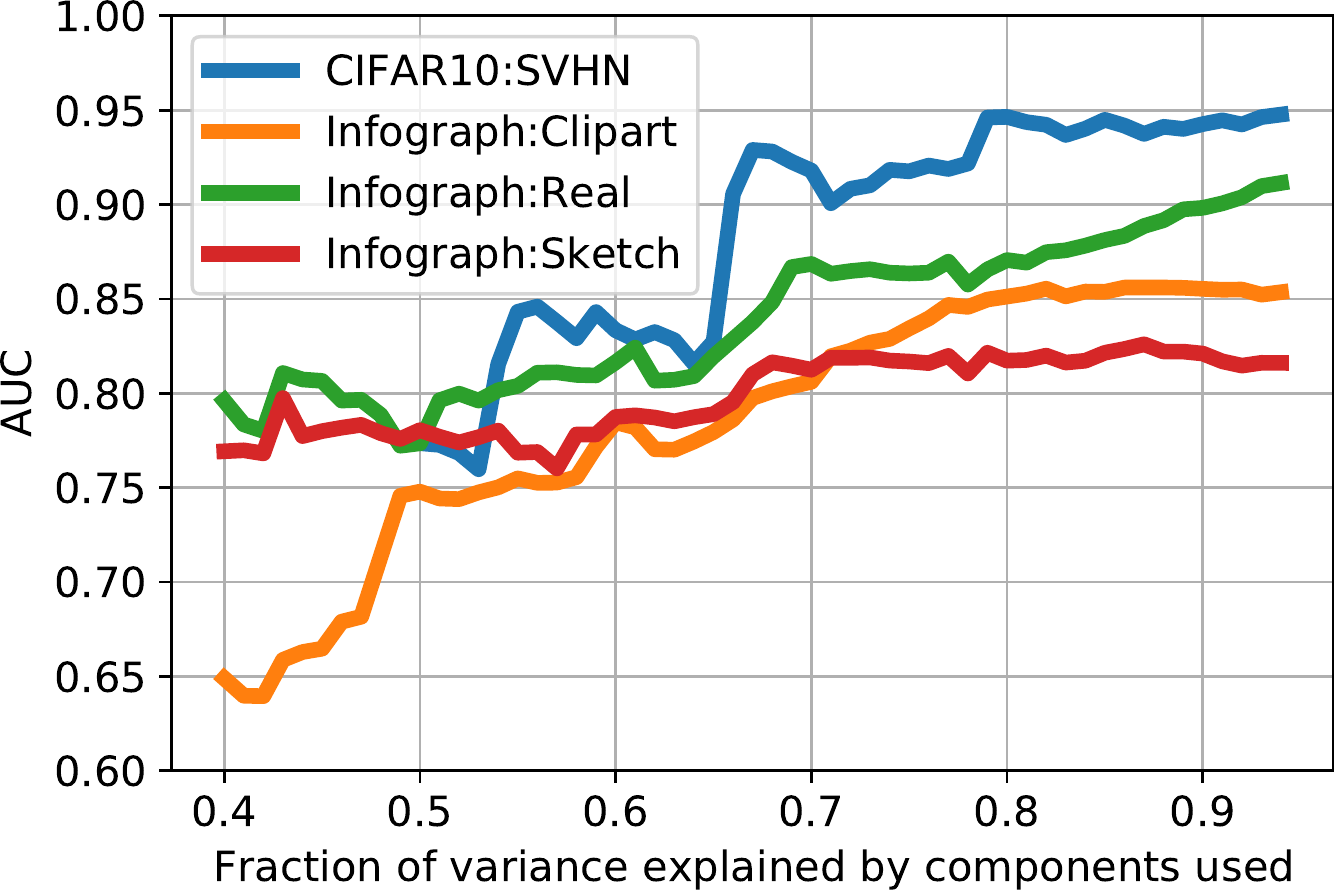} &
    \includegraphics[width=0.33\linewidth]{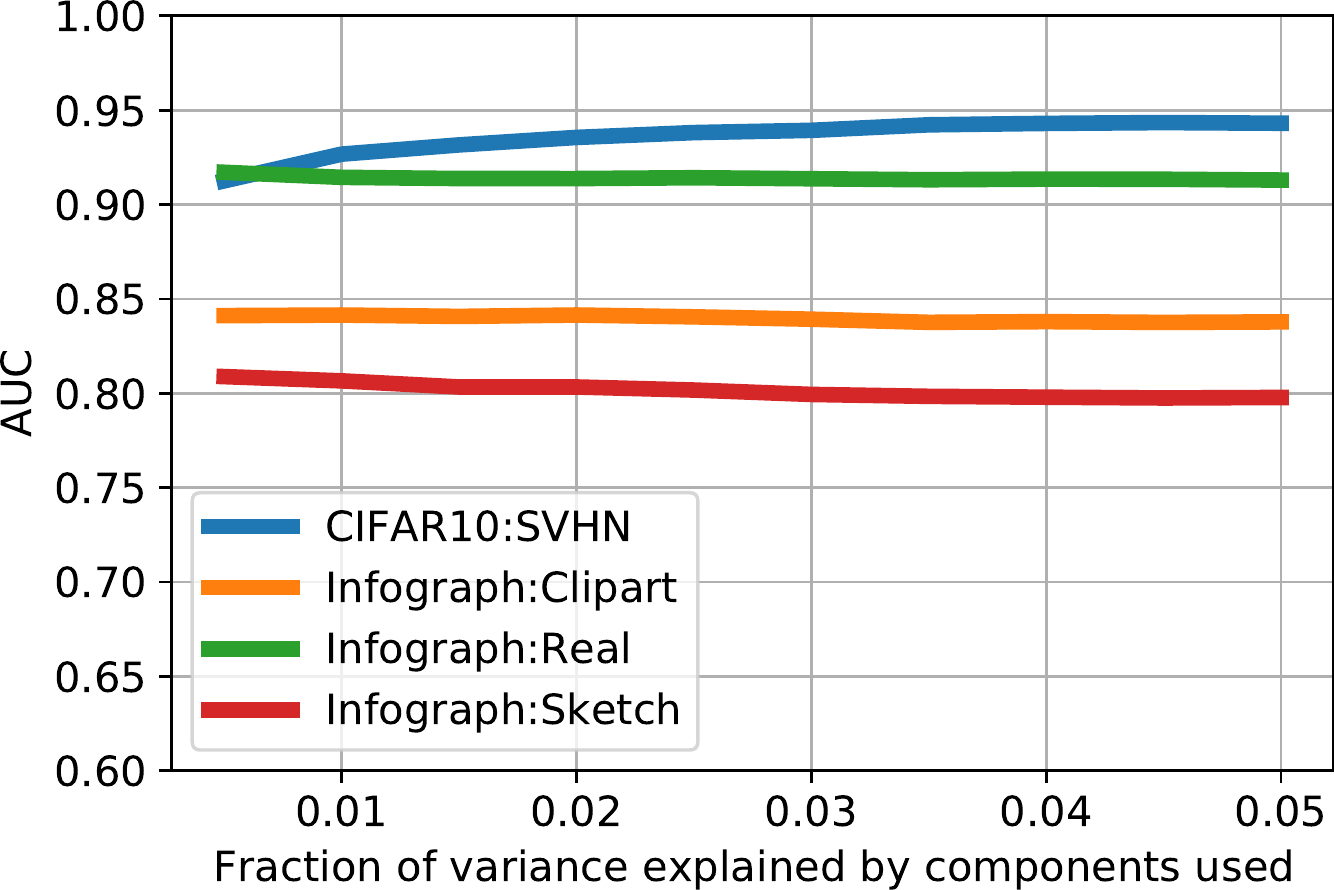} \\
    (a) & (b)\\
  \end{tabular}
    \caption{
    The OOD AUC for four experiments with a ResNet50 using different sets of principal components. (a)~gives result starting from the first principal components, while (b)~does so from the last principal components. The x-axis of (a) starts at~0.4 as for Infograph the most variant component of the first layer is responsible for almost~40\% of all variance.}
    \label{fig:auc_per_component}
\end{figure}

In order to assess the importance of data invariants for U-OOD detection, we examined which principal components are most effective at identifying OOD samples. To that end, we measured the AUC score in four experiments by limiting the Mahalanobis score of Eq.~\eqref{eq:score_g} to only use the subset of principal components with highest variance, corresponding to the modes of variation of the data. Similarly, we observed the performance with the subset of principal components with the smallest variance, corresponding to data invariants. The latter outperformed the former by a large margin in U-OOD detection (see Fig.~\ref{fig:auc_per_component}). Starting from the most variant principal components, the performance slowly increases when adding more components, converging when over~80\% of the variance is explained. On the other hand, when starting from the most invariant component, the performance quickly converges when as little as~$3\%$ of the variance has been explained, supporting the idea that invariants are more representative to characterize training data and OOD~samples. While  other works had observed similar findings, they either consider the supervised case~\cite{kamoi2020mahalanobis,podolskiy2021revisiting}, or frame it in the context of reducing dimensionality~\cite{rippel2021modeling}.

An interesting consequence of our invariant-based interpretation of U-OOD is that, when we considered what to include in our benchmark, some experiments that are valid for evaluating supervised OOD methods, are in fact not suitable for the U-OOD case. For instance, CIFAR10:CIFAR100~\cite{mohseni2021multi,morningstar2021density,sehwag2021ssd} or 9-classes:1-held-out-class of CIFAR10~\cite{akcay2018ganomaly,bergman2020deep} were used in previous U-OOD works even though they do not appear to meet the U-OOD criteria.

More specifically, according to our definition for U-OOD, one would expect that by increasing the number of classes present in a training set, the invariants associated to the high semantic features will decrease, effectively reducing the probability that new classes are considered~U-OOD. For example, training with multiple classes from CIFAR10 (\emph{e.g.},~cats, dogs, cars) reduces the probability that an additional class (\emph{e.g.},~plane) from CIFAR10 or CIFAR100 should be considered~OOD (Fig.~\ref{fig:geometric}(d)), as the class stops being an invariant. However, the number of training classes should not affect the probability that images from a different modality are detected as~OOD, as they break other kind of invariants. For instance, when training with images from CIFAR10, the test images from SVHN or MNIST should still be considered~OOD regardless of the number of CIFAR10 training classes, as they are clearly distinct in appearance. 

\begin{figure}[t]
  \centering
  \setlength\tabcolsep{16pt}
  \begin{tabular}{cc}
    \includegraphics[width=0.33\linewidth]{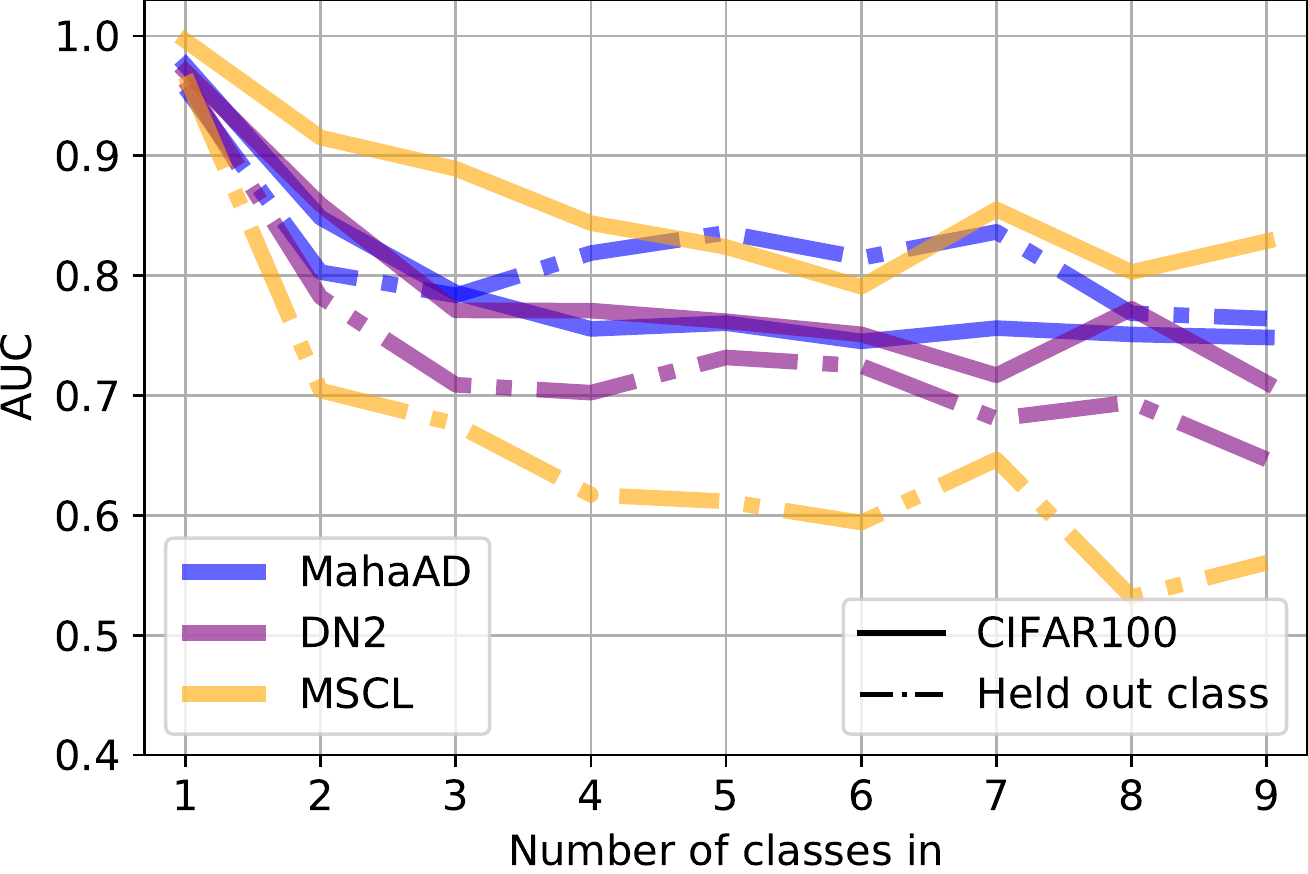} &
    \includegraphics[width=0.33\linewidth]{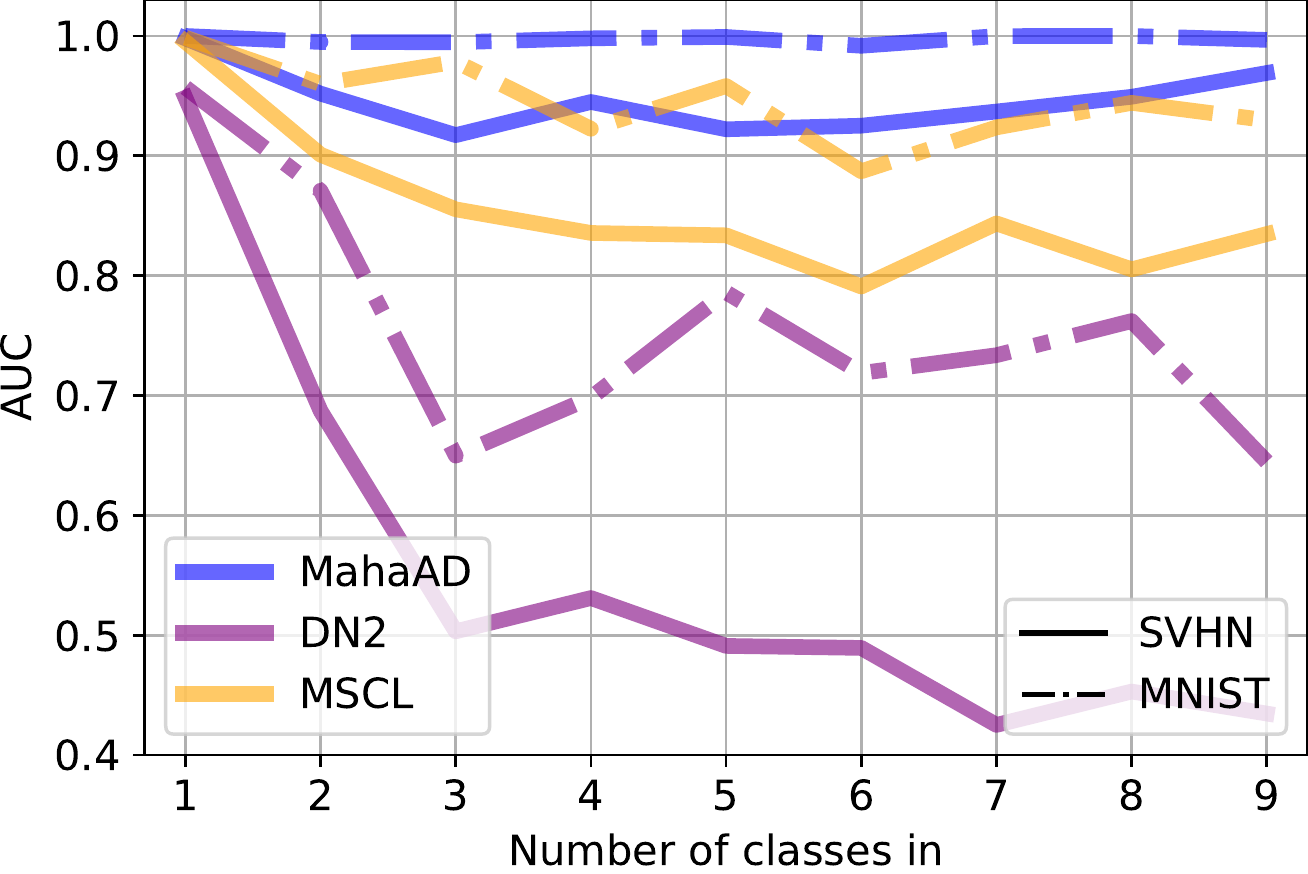} \\
    (a) & (b)\\
  \end{tabular}
    \caption{OOD AUC performance for different methods as a function of the number of classes in the training set (CIFAR10), keeping its size constant. (a)~Performance on CIFAR100 and a held out CIFAR10 class as out-distribution, which should not be considered U-OOD under our interpretation. (b)~Performance on SVHN and a held out MNIST class as out-distribution, which should be considered U-OOD under our interpretation.
    }
    \label{fig:vary_classes}
\end{figure}

In Fig.~\ref{fig:vary_classes}, we investigate this desired behavior experimentally by analyzing the performance of the three best-performing methods when we increased the number of in-distribution CIFAR10 training classes. As expected, all methods consider fewer images from CIFAR100 and one held-out class from CIFAR10 as OOD when the number of training classes increased~(Fig.~\ref{fig:vary_classes}(a)). Conversely, increasing the number of training CIFAR10 classes did not affect the predictions for SVHN and MNIST with~\MahaAD, which correctly kept detecting both datasets as~OOD~(Fig.~\ref{fig:vary_classes}(b)). In contrast, this did negatively affect the predictions of \textbf{DN2} and \MSCL{}. According to our invariant-based interpretation of U-OOD, \MahaAD's behavior is reasonable and consistent in these configurations, yet the unexpected \textbf{DN2} and \MSCL{} results are hard to justify. To the extent of our knowledge, no previous work on U-OOD detection had provided a similar theoretical tool capable of interpreting and explaining results.

\section{Conclusion}\label{sec:conc}
Our work explores the state of U-OOD~detection by observing the behavior of methods on an extensive and varied set of tasks. By doing so, we show a complicated landscape, with most methods being highly inconsistent among and within tasks. \MahaAD~is however an exception to this trend, behaving consistently in a large majority of experimental configurations. Despite being neglected in most recent U-OOD papers, \MahaAD~appears to be the current best off-the-shelf unsupervised OOD detector, as it offers good performance and consistency without requiring time-consuming data pre-processing, careful tuning of the training procedure, or hyperparameter search.

In order to explain these inconsistent results, we introduced a characterization of U-OOD based on training set invariants and showed that the \MahaAD~method embodies a linear version of this concept.
We found this framework and the proposed benchmark to be useful to not only qualitatively understand U-OOD detector predictions, but also to assess whether a test dataset is in fact {suitable for U-OOD evaluation} or not. A key take-away is that we cannot purely rely on semantic labels from datasets to design U-OOD evaluation methods, as done in previous works.

In general, this points to a rather bleak conclusion: at the moment, no method can consistently outperform a simple anomaly detector that uses naively extracted features from a network trained on a different dataset that was optimized for a different task. We believe that with our invariant-based U-OOD characterization, new appropriate methods can be designed and validated in comprehensive ways.

\subsection*{Acknowledgements} This work was funded by the Swiss National Science Foundation (SNSF), research grant 200021\_192285 “Image data validation for AI systems”.

\clearpage
%
%
\bibliographystyle{ECCV/splncs04}
\bibliography{refs}

\begin{thebibliography}{10}
\providecommand{\url}[1]{\texttt{#1}}
\providecommand{\urlprefix}{URL }
\providecommand{\doi}[1]{https://doi.org/#1}

\bibitem{abati2019latent}
Abati, D., Porrello, A., Calderara, S., Cucchiara, R.: Latent space
  autoregression for novelty detection. In: Proceedings of the IEEE/CVF
  Conference on Computer Vision and Pattern Recognition. pp. 481--490 (2019)

\bibitem{ahmed2020detecting}
Ahmed, F., Courville, A.: Detecting semantic anomalies. In: Proceedings of the
  AAAI Conference on Artificial Intelligence. vol.~34, pp. 3154--3162 (2020)

\bibitem{akcay2018ganomaly}
Akcay, S., Atapour-Abarghouei, A., Breckon, T.P.: Ganomaly: Semi-supervised
  anomaly detection via adversarial training. In: Asian conference on computer
  vision. pp. 622--637. Springer (2018)

\bibitem{amodei2016concrete}
Amodei, D., Olah, C., Steinhardt, J., Christiano, P., Schulman, J., Man{\'e},
  D.: Concrete problems in ai safety. arXiv preprint arXiv:1606.06565  (2016)

\bibitem{battikh2021latent}
Battikh, M.S., Lenskiy, A.A.: Latent-insensitive autoencoders for anomaly
  detection and class-incremental learning. arXiv preprint arXiv:2110.13101
  (2021)

\bibitem{bergman2020deep}
Bergman, L., Cohen, N., Hoshen, Y.: Deep nearest neighbor anomaly detection.
  arXiv preprint arXiv:2002.10445  (2020)

\bibitem{bergman2020classification}
Bergman, L., Hoshen, Y.: Classification-based anomaly detection for general
  data. International Conference on Learning Representations  (2020)

\bibitem{Bergmann_2019_CVPR}
Bergmann, P., Fauser, M., Sattlegger, D., Steger, C.: Mvtec ad -- a
  comprehensive real-world dataset for unsupervised anomaly detection. In:
  Proceedings of the IEEE/CVF Conference on Computer Vision and Pattern
  Recognition (CVPR) (6 2019)

\bibitem{bergmann2020uninformed}
Bergmann, P., Fauser, M., Sattlegger, D., Steger, C.: Uninformed students:
  Student-teacher anomaly detection with discriminative latent embeddings. In:
  Proceedings of the IEEE/CVF Conference on Computer Vision and Pattern
  Recognition. pp. 4183--4192 (2020)

\bibitem{bergmann2018improving}
Bergmann, P., L{\"o}we, S., Fauser, M., Sattlegger, D., Steger, C.: Improving
  unsupervised defect segmentation by applying structural similarity to
  autoencoders. arXiv preprint arXiv:1807.02011  (2018)

\bibitem{bozorgtabar2020salad}
Bozorgtabar, B., Mahapatra, D., Vray, G., Thiran, J.P.: Salad: Self-supervised
  aggregation learning for anomaly detection on x-rays. In: International
  Conference on Medical Image Computing and Computer-Assisted Intervention. pp.
  468--478. Springer (2020)

\bibitem{breunig2000lof}
Breunig, M.M., Kriegel, H.P., Ng, R.T., Sander, J.: Lof: identifying
  density-based local outliers. In: Proceedings of the 2000 ACM SIGMOD
  international conference on Management of data. pp. 93--104 (2000)

\bibitem{chalapathy2019deep}
Chalapathy, R., Chawla, S.: Deep learning for anomaly detection: A survey.
  arXiv preprint arXiv:1901.03407  (2019)

\bibitem{chalapathy2018anomaly}
Chalapathy, R., Menon, A.K., Chawla, S.: Anomaly detection using one-class
  neural networks. arXiv preprint arXiv:1802.06360  (2018)

\bibitem{choi2018waic}
Choi, H., Jang, E., Alemi, A.A.: Waic, but why? generative ensembles for robust
  anomaly detection. arXiv preprint arXiv:1810.01392  (2018)

\bibitem{choi2019novelty}
Choi, S., Chung, S.Y.: Novelty detection via blurring. International Conference
  on Learning Representations  (2020)

\bibitem{cohen2020sub}
Cohen, N., Hoshen, Y.: Sub-image anomaly detection with deep pyramid
  correspondences. arXiv preprint arXiv:2005.02357  (2020)

\bibitem{collin2021improved}
Collin, A.S., De~Vleeschouwer, C.: Improved anomaly detection by training an
  autoencoder with skip connections on images corrupted with stain-shaped
  noise. In: 2020 25th International Conference on Pattern Recognition (ICPR).
  pp. 7915--7922. IEEE (2021)

\bibitem{defard2021padim}
Defard, T., Setkov, A., Loesch, A., Audigier, R.: Padim: a patch distribution
  modeling framework for anomaly detection and localization. In: International
  Conference on Pattern Recognition. pp. 475--489. Springer (2021)

\bibitem{dehaene2020iterative}
Dehaene, D., Frigo, O., Combrexelle, S., Eline, P.: Iterative energy-based
  projection on a normal data manifold for anomaly localization. arXiv preprint
  arXiv:2002.03734  (2020)

\bibitem{du2019implicit}
Du, Y., Mordatch, I.: Implicit generation and generalization in energy-based
  models. arXiv preprint arXiv:1903.08689  (2019)

\bibitem{golan2018deep}
Golan, I., El-Yaniv, R.: Deep anomaly detection using geometric
  transformations. In: Advances in Neural Information Processing Systems. pp.
  9758--9769 (2018)

\bibitem{gong2019memorizing}
Gong, D., Liu, L., Le, V., Saha, B., Mansour, M.R., Venkatesh, S., Hengel,
  A.v.d.: Memorizing normality to detect anomaly: Memory-augmented deep
  autoencoder for unsupervised anomaly detection. In: Proceedings of the
  IEEE/CVF International Conference on Computer Vision. pp. 1705--1714 (2019)

\bibitem{goyal2020drocc}
Goyal, S., Raghunathan, A., Jain, M., Simhadri, H.V., Jain, P.: Drocc: Deep
  robust one-class classification. In: International Conference on Machine
  Learning. pp. 3711--3721. PMLR (2020)

\bibitem{graham2015kaggle}
Graham, B.: Kaggle diabetic retinopathy detection competition report.
  University of Warwick  (2015)

\bibitem{gudovskiy2022cflow}
Gudovskiy, D., Ishizaka, S., Kozuka, K.: Cflow-ad: Real-time unsupervised
  anomaly detection with localization via conditional normalizing flows. In:
  Proceedings of the IEEE/CVF Winter Conference on Applications of Computer
  Vision. pp. 98--107 (2022)

\bibitem{havtorn2021hierarchical}
Havtorn, J.D.D., Frellsen, J., Hauberg, S., Maal{\o}e, L.: Hierarchical vaes
  know what they don’t know. In: International Conference on Machine
  Learning. pp. 4117--4128. PMLR (2021)

\bibitem{hendrycks2016baseline}
Hendrycks, D., Gimpel, K.: A baseline for detecting misclassified and
  out-of-distribution examples in neural networks. International Conference on
  Learning Representations  (2017)

\bibitem{hendrycks2019using}
Hendrycks, D., Mazeika, M., Kadavath, S., Song, D.: Using self-supervised
  learning can improve model robustness and uncertainty. Advances in Neural
  Information Processing Systems  \textbf{32} (2019)

\bibitem{hjelm2018learning}
Hjelm, R.D., Fedorov, A., Lavoie-Marchildon, S., Grewal, K., Bachman, P.,
  Trischler, A., Bengio, Y.: Learning deep representations by mutual
  information estimation and maximization. arXiv preprint arXiv:1808.06670
  (2018)

\bibitem{hou2021divide}
Hou, J., Zhang, Y., Zhong, Q., Xie, D., Pu, S., Zhou, H.: Divide-and-assemble:
  Learning block-wise memory for unsupervised anomaly detection. In:
  Proceedings of the IEEE/CVF International Conference on Computer Vision. pp.
  8791--8800 (2021)

\bibitem{hsu2020generalized}
Hsu, Y.C., Shen, Y., Jin, H., Kira, Z.: Generalized odin: Detecting
  out-of-distribution image without learning from out-of-distribution data. In:
  Proceedings of the IEEE/CVF Conference on Computer Vision and Pattern
  Recognition. pp. 10951--10960 (2020)

\bibitem{huang2019attribute}
Huang, C., Ye, F., Cao, J., Li, M., Zhang, Y., Lu, C.: Attribute restoration
  framework for anomaly detection. arXiv preprint arXiv:1911.10676  (2019)

\bibitem{ji2018invariant}
Ji, X., Henriques, J.F., Vedaldi, A.: Invariant information distillation for
  unsupervised image segmentation and clustering. arXiv preprint
  arXiv:1807.06653  \textbf{2}(3), ~8 (2018)

\bibitem{kamoi2020mahalanobis}
Kamoi, R., Kobayashi, K.: Why is the mahalanobis distance effective for anomaly
  detection? arXiv preprint arXiv:2003.00402  (2020)

\bibitem{kingma2018glow}
Kingma, D.P., Dhariwal, P.: Glow: Generative flow with invertible 1x1
  convolutions. Advances in neural information processing systems  \textbf{31}
  (2018)

\bibitem{koner2021oodformer}
Koner, R., Sinhamahapatra, P., Roscher, K., G{\"u}nnemann, S., Tresp, V.:
  Oodformer: Out-of-distribution detection transformer. arXiv preprint
  arXiv:2107.08976  (2021)

\bibitem{krizhevsky2009learning}
Krizhevsky, A., Hinton, G., et~al.: Learning multiple layers of features from
  tiny images  (2009)

\bibitem{ledoit2004well}
Ledoit, O., Wolf, M.: A well-conditioned estimator for large-dimensional
  covariance matrices. Journal of multivariate analysis  \textbf{88}(2),
  365--411 (2004)

\bibitem{lee2018simple}
Lee, K., Lee, K., Lee, H., Shin, J.: A simple unified framework for detecting
  out-of-distribution samples and adversarial attacks. Advances in Neural
  Information Processing Systems  \textbf{31},  7167--7177 (2018)

\bibitem{li2021cutpaste}
Li, C.L., Sohn, K., Yoon, J., Pfister, T.: Cutpaste: Self-supervised learning
  for anomaly detection and localization. In: Proceedings of the IEEE/CVF
  Conference on Computer Vision and Pattern Recognition. pp. 9664--9674 (2021)

\bibitem{li2019exploring}
Li, X., Kiringa, I., Yeap, T., Zhu, X., Li, Y.: Exploring deep anomaly
  detection methods based on capsule net. arXiv preprint arXiv:1907.06312
  (2019)

\bibitem{liang2017enhancing}
Liang, S., Li, Y., Srikant, R.: Enhancing the reliability of
  out-of-distribution image detection in neural networks. International
  Conference on Learning Representations  (2018)

\bibitem{liu2008isolation}
Liu, F.T., Ting, K.M., Zhou, Z.H.: Isolation forest. In: 2008 eighth ieee
  international conference on data mining. pp. 413--422. IEEE (2008)

\bibitem{marquez2019image}
M{\'a}rquez-Neila, P., Sznitman, R.: Image data validation for medical systems.
  In: International Conference on Medical Image Computing and Computer-Assisted
  Intervention. pp. 329--337. Springer (2019)

\bibitem{mesarcik2021improving}
Mesarcik, M., Ranguelova, E., Boonstra, A.J., van Nieuwpoort, R.V.: Improving
  novelty detection using the reconstructions of nearest neighbours. arXiv
  preprint arXiv:2111.06150  (2021)

\bibitem{mohseni2021multi}
Mohseni, S., Vahdat, A., Yadawa, J.: Multi-task transformation learning for
  robust out-of-distribution detection. arXiv preprint arXiv:2106.03899  (2021)

\bibitem{morningstar2021density}
Morningstar, W., Ham, C., Gallagher, A., Lakshminarayanan, B., Alemi, A.,
  Dillon, J.: Density of states estimation for out of distribution detection.
  In: International Conference on Artificial Intelligence and Statistics. pp.
  3232--3240. PMLR (2021)

\bibitem{nalisnick2018deep}
Nalisnick, E., Matsukawa, A., Teh, Y.W., Gorur, D., Lakshminarayanan, B.: Do
  deep generative models know what they don't know? arXiv preprint
  arXiv:1810.09136  (2018)

\bibitem{nalisnick2019detecting}
Nalisnick, E., Matsukawa, A., Teh, Y.W., Lakshminarayanan, B.: Detecting
  out-of-distribution inputs to deep generative models using a test for
  typicality. International Conference on Learning Representations  (2019)

\bibitem{netzer2011reading}
Netzer, Y., Wang, T., Coates, A., Bissacco, A., Wu, B., Ng, A.Y.: Reading
  digits in natural images with unsupervised feature learning  (2011)

\bibitem{ouardini2019towards}
Ouardini, K., Yang, H., Unnikrishnan, B., Romain, M., Garcin, C., Zenati, H.,
  Campbell, J.P., Chiang, M.F., Kalpathy-Cramer, J., Chandrasekhar, V., et~al.:
  Towards practical unsupervised anomaly detection on retinal images. In:
  Domain Adaptation and Representation Transfer and Medical Image Learning with
  Less Labels and Imperfect Data, pp. 225--234. Springer (2019)

\bibitem{peng2019moment}
Peng, X., Bai, Q., Xia, X., Huang, Z., Saenko, K., Wang, B.: Moment matching
  for multi-source domain adaptation. In: Proceedings of the IEEE/CVF
  International Conference on Computer Vision. pp. 1406--1415 (2019)

\bibitem{perera2019ocgan}
Perera, P., Nallapati, R., Xiang, B.: Ocgan: One-class novelty detection using
  gans with constrained latent representations. In: Proceedings of the IEEE/CVF
  Conference on Computer Vision and Pattern Recognition. pp. 2898--2906 (2019)

\bibitem{perera2021one}
Perera, P., Oza, P., Patel, V.M.: One-class classification: A survey. arXiv
  preprint arXiv:2101.03064  (2021)

\bibitem{podolskiy2021revisiting}
Podolskiy, A., Lipin, D., Bout, A., Artemova, E., Piontkovskaya, I.: Revisiting
  mahalanobis distance for transformer-based out-of-domain detection. arXiv
  preprint arXiv:2101.03778  (2021)

\bibitem{reiss2021panda}
Reiss, T., Cohen, N., Bergman, L., Hoshen, Y.: Panda: Adapting pretrained
  features for anomaly detection and segmentation. In: Proceedings of the
  IEEE/CVF Conference on Computer Vision and Pattern Recognition. pp.
  2806--2814 (2021)

\bibitem{reiss2021mean}
Reiss, T., Hoshen, Y.: Mean-shifted contrastive loss for anomaly detection.
  arXiv preprint arXiv:2106.03844  (2021)

\bibitem{ren2019likelihood}
Ren, J., Liu, P.J., Fertig, E., Snoek, J., Poplin, R., Depristo, M., Dillon,
  J., Lakshminarayanan, B.: Likelihood ratios for out-of-distribution
  detection. In: Advances in Neural Information Processing Systems. pp.
  14707--14718 (2019)

\bibitem{rippel2021modeling}
Rippel, O., Mertens, P., Merhof, D.: Modeling the distribution of normal data
  in pre-trained deep features for anomaly detection. In: 2020 25th
  International Conference on Pattern Recognition (ICPR). pp. 6726--6733. IEEE
  (2021)

\bibitem{ruff2021unifying}
Ruff, L., Kauffmann, J.R., Vandermeulen, R.A., Montavon, G., Samek, W., Kloft,
  M., Dietterich, T.G., M{\"u}ller, K.R.: A unifying review of deep and shallow
  anomaly detection. Proceedings of the IEEE  (2021)

\bibitem{ruff2018deep}
Ruff, L., Vandermeulen, R., Goernitz, N., Deecke, L., Siddiqui, S.A., Binder,
  A., M{\"u}ller, E., Kloft, M.: Deep one-class classification. In:
  International conference on machine learning. pp. 4393--4402. PMLR (2018)

\bibitem{sabokrou2018avid}
Sabokrou, M., Pourreza, M., Fayyaz, M., Entezari, R., Fathy, M., Gall, J.,
  Adeli, E.: Avid: Adversarial visual irregularity detection. In: Asian
  Conference on Computer Vision. pp. 488--505. Springer (2018)

\bibitem{sakurada2014anomaly}
Sakurada, M., Yairi, T.: Anomaly detection using autoencoders with nonlinear
  dimensionality reduction. In: Proceedings of the MLSDA 2014 2nd Workshop on
  Machine Learning for Sensory Data Analysis. pp. 4--11 (2014)

\bibitem{salehi2020arae}
Salehi, M., Arya, A., Pajoum, B., Otoofi, M., Shaeiri, A., Rohban, M.H.,
  Rabiee, H.R.: Arae: Adversarially robust training of autoencoders improves
  novelty detection. arXiv preprint arXiv:2003.05669  (2020)

\bibitem{salehi2020puzzle}
Salehi, M., Eftekhar, A., Sadjadi, N., Rohban, M.H., Rabiee, H.R.: Puzzle-ae:
  Novelty detection in images through solving puzzles. arXiv preprint
  arXiv:2008.12959  (2020)

\bibitem{salehi2021multiresolution}
Salehi, M., Sadjadi, N., Baselizadeh, S., Rohban, M.H., Rabiee, H.R.:
  Multiresolution knowledge distillation for anomaly detection. In: Proceedings
  of the IEEE/CVF Conference on Computer Vision and Pattern Recognition. pp.
  14902--14912 (2021)

\bibitem{salimans2017pixelcnn++}
Salimans, T., Karpathy, A., Chen, X., Kingma, D.P.: Pixelcnn++: Improving the
  pixelcnn with discretized logistic mixture likelihood and other
  modifications. International Conference on Learning Representations  (2017)

\bibitem{sastry2020detecting}
Sastry, C.S., Oore, S.: Detecting out-of-distribution examples with gram
  matrices. In: International Conference on Machine Learning. pp. 8491--8501.
  PMLR (2020)

\bibitem{schirrmeister2020understanding}
Schirrmeister, R., Zhou, Y., Ball, T., Zhang, D.: Understanding anomaly
  detection with deep invertible networks through hierarchies of distributions
  and features. Advances in Neural Information Processing Systems  \textbf{33},
   21038--21049 (2020)

\bibitem{schlegl2017unsupervised}
Schlegl, T., Seeb{\"o}ck, P., Waldstein, S.M., Schmidt-Erfurth, U., Langs, G.:
  Unsupervised anomaly detection with generative adversarial networks to guide
  marker discovery. In: International conference on information processing in
  medical imaging. pp. 146--157. Springer (2017)

\bibitem{scholkopf1999support}
Sch{\"o}lkopf, B., Williamson, R.C., Smola, A.J., Shawe-Taylor, J., Platt,
  J.C., et~al.: Support vector method for novelty detection. In: NIPS. vol.~12,
  pp. 582--588. Citeseer (1999)

\bibitem{sehwag2021ssd}
Sehwag, V., Chiang, M., Mittal, P.: Ssd: A unified framework for
  self-supervised outlier detection. International Conference on Learning
  Representations  (2021)

\bibitem{serra2019input}
Serr{\`a}, J., {\'A}lvarez, D., G{\'o}mez, V., Slizovskaia, O., N{\'u}{\~n}ez,
  J.F., Luque, J.: Input complexity and out-of-distribution detection with
  likelihood-based generative models. International Conference on Learning
  Representations  (2019)

\bibitem{sohn2020learning}
Sohn, K., Li, C.L., Yoon, J., Jin, M., Pfister, T.: Learning and evaluating
  representations for deep one-class classification. International Conference
  on Learning Representations  (2021)

\bibitem{tack2020csi}
Tack, J., Mo, S., Jeong, J., Shin, J.: Csi: Novelty detection via contrastive
  learning on distributionally shifted instances. Advances in neural
  information processing systems  \textbf{33},  11839--11852 (2020)

\bibitem{tang2019abnormal}
Tang, Y.X., Tang, Y.B., Han, M., Xiao, J., Summers, R.M.: Abnormal chest x-ray
  identification with generative adversarial one-class classifier. In: 2019
  IEEE 16th International Symposium on Biomedical Imaging (ISBI 2019). pp.
  1358--1361. IEEE (2019)

\bibitem{tang2020automated}
Tang, Y.X., Tang, Y.B., Peng, Y., Yan, K., Bagheri, M., Redd, B.A., Brandon,
  C.J., Lu, Z., Han, M., Xiao, J., et~al.: Automated abnormality classification
  of chest radiographs using deep convolutional neural networks. NPJ digital
  medicine  \textbf{3}(1), ~1--8 (2020)

\bibitem{torralba200880}
Torralba, A., Fergus, R., Freeman, W.T.: 80 million tiny images: A large data
  set for nonparametric object and scene recognition. IEEE transactions on
  pattern analysis and machine intelligence  \textbf{30}(11),  1958--1970
  (2008)

\bibitem{venkataramanan2020attention}
Venkataramanan, S., Peng, K.C., Singh, R.V., Mahalanobis, A.: Attention guided
  anomaly localization in images. In: European Conference on Computer Vision.
  pp. 485--503. Springer (2020)

\bibitem{wang2019effective}
Wang, S., Zeng, Y., Liu, X., Zhu, E., Yin, J., Xu, C., Kloft, M.: Effective
  end-to-end unsupervised outlier detection via inlier priority of
  discriminative network. In: Advances in Neural Information Processing
  Systems. pp. 5962--5975 (2019)

\bibitem{wang2017chestx}
Wang, X., Peng, Y., Lu, L., Lu, Z., Bagheri, M., Summers, R.M.: Chestx-ray8:
  Hospital-scale chest x-ray database and benchmarks on weakly-supervised
  classification and localization of common thorax diseases. In: Proceedings of
  the IEEE conference on computer vision and pattern recognition. pp.
  2097--2106 (2017)

\bibitem{xiao2020vaebm}
Xiao, Z., Kreis, K., Kautz, J., Vahdat, A.: Vaebm: A symbiosis between
  variational autoencoders and energy-based models. In: International
  Conference on Learning Representations (2020)

\bibitem{xiao2020likelihood}
Xiao, Z., Yan, Q., Amit, Y.: Likelihood regret: An out-of-distribution
  detection score for variational auto-encoder. arXiv preprint arXiv:2003.02977
   (2020)

\bibitem{xiao2021we}
Xiao, Z., Yan, Q., Amit, Y.: Do we really need to learn representations from
  in-domain data for outlier detection? ICML 2021 Workshop on Uncertainty \&
  Robustness in Deep Learning  (2021)

\bibitem{yang2021generalized}
Yang, J., Zhou, K., Li, Y., Liu, Z.: Generalized out-of-distribution detection:
  A survey. arXiv preprint arXiv:2110.11334  (2021)

\bibitem{yoon2021autoencoding}
Yoon, S., Noh, Y.K., Park, F.C.: Autoencoding under normalization constraints.
  arXiv preprint arXiv:2105.05735  (2021)

\bibitem{zhai2016deep}
Zhai, S., Cheng, Y., Lu, W., Zhang, Z.: Deep structured energy based models for
  anomaly detection. In: International Conference on Machine Learning. pp.
  1100--1109. PMLR (2016)

\bibitem{zong2018deep}
Zong, B., Song, Q., Min, M.R., Cheng, W., Lumezanu, C., Cho, D., Chen, H.: Deep
  autoencoding gaussian mixture model for unsupervised anomaly detection. In:
  International conference on learning representations (2018)

\end{thebibliography}

\section{Supplementary Material}
\subsection{Dataset Details}
We briefly describe all datasets used in our experiments. An overview of our experimental set-up is given in Table~\ref{tab:exps}.

\begin{description}
    \item [CIFAR10~\cite{krizhevsky2009learning}.] (In)~Small, natural images divided into 10 classes. For \uniclass, one class forms the in-distribution, with its test set used in the evaluation. For \shiftlowres, all 50000 training images are used for training when considered in-distribution, and all 10000 test images are used for testing. (Out)~The remaining 9 classes are used as OOD for \uniclass, subsampled to 1000~images.
    \item [CIFAR100~\cite{krizhevsky2009learning}.] (In)~20 experiments with the training set of one of the semantic superclasses as the in-distribution, with its test set used during evaluation. (Out)~Images from the remaining superclasses, subsampled to 500~images.
    \item [MVTec~\cite{Bergmann_2019_CVPR}.] (In)~Between 60 and 391 aligned images of 15~different objects and textures. 12-60 images are used as the in-distribution at test time. (Out)~30-141 images of defect objects are used as OOD.
    \item[OCT.] (In)~A collection of 58849~retinal Optical Coherence Tomography images used for training, and 300 for testing. (Out)~Corrupted OCT scans built as described in~\cite{marquez2019image}.
    \item[Chest~\cite{wang2017chestx}.] (In)~The NIH Clinical Center ChestX-ray dataset containing 85524 training images. We use 300 images from the test set during evaluation. (Out)~Corrupted X-ray scans as described in~\cite{marquez2019image}.
    \item[NIH~\cite{tang2020automated}.] (In)~A collection of 4261~healthy X-ray scans of the NIH Clinical Center ChestX-ray dataset. The healthy test scans are used as the in-distribution during evaluation. (Out)~Pathological scans from the same dataset.
    \item[DRD~\cite{graham2015kaggle}.] (In)~25809~healthy high-resolution retinal fundus photographs. Healthy test scans are again used during evaluation.\\ (Out)~Retinal fundus photographs depicting 4~different levels of diabetic retinopathy (DR). The level of~DR is indicated by a digit next to the method's name~(DRD1--DRD4).
    \item [SVHN~\cite{netzer2011reading}.] A dataset consisting of images of house numbers. We only use it as an OOD dataset, where the test set is reduced to 10000~samples.
    \item [DomainNet~\cite{peng2019moment}.] (In)~The train and test images from the first 173 classes are used for training and evaluation respectively (as in \cite{hsu2020generalized}). We perform 10 experiments with the real images, and 10 with infographs. (Out)~10 domain-class combinations are used as OOD datasets. We avoid using \texttt{Real-B} and \texttt{Infograph-B} as OOD in the first and the second group of experiments respectively. All test sets are downsampled to 5000~images.
\end{description}

\begin{table*}
\caption{Experimental set-up. }
\label{tab:exps}
\begin{center}
\resizebox{\textwidth}{!}{
\begin{tabular}{cccccc}
\hline
 Category& \# Tasks & Tasks & \# train & \# in & \# out \\
\hline 
\multirow{8}{*}{\uniclass} & \multirow{2}{*}{10}&  \{airplane,automobile,bird,cat,deer,  &\multirow{2}{*}{5000} & \multirow{2}{*}{1000} & \multirow{2}{*}{1000}\\
&&dog,frog,horse,ship,truck\}:rest&&\\
 & \multirow{6}{*}{20}&  \{aquatic mammals,fish,flowers,food containers,fruit and vegetables, &\multirow{6}{*}{2500} & \multirow{6}{*}{500} & \multirow{6}{*}{500}\\
&&household electrical devices,household furniture,insects,&&\\
&&large carnivores,large man-made outdoor things,&&\\
&&large natural outdoor scenes,large omnivores and herbivores,&&\\
&&medium-sized mammals,non-insect invertebrates,&&\\
&&people,reptiles,small mammals,trees,vehicles 1,vehicles 2\}:rest&&\\
\hline
\multirow{3}{*}{\uniano} & \multirow{3}{*}{15} & \{bottle,cable,capsule,carpet,grid,hazelnut, & \multirow{3}{*}{60-391} & \multirow{3}{*}{12-60} & \multirow{3}{*}{30-141}\\
&&leather,metal nut,pill,screw,tile,&&\\
&&toothbrush,transistor,wood,zipper\}:defect&&\\
\hline
\multirow{4}{*}{\unimed} &1 & OCT:corruptions & 58849 & 300 & 300\\
 & 1& Chest:corruptions & 85524 & 300 & 300\\
 & 1& NIH:pathology & 4261 & 677 & 667\\
 & 4& DRD:DRD1-4 & 25809 & 500& 500\\
\hline
\shiftlowres & 1 & CIFAR10:SVHN & 50000 & 10000 & 10000\\
\hline
\multirow{6}{*}{\shifthighres} &\multirow{3}{*}{10}& Real A:\{Quickdraw A,Quickdraw B,Infograph A, & \multirow{3}{*}{61817} &\multirow{3}{*}{5000} & \multirow{3}{*}{5000}\\
&&Infograph B,Sketch A,Sketch B,&&\\
&&Clipart A,Clipart B,Painting A,Painting B\}&&\\

&\multirow{3}{*}{10}& Infograph A:\{Quickdraw A,Quickdraw B, & \multirow{3}{*}{14069} &\multirow{3}{*}{5000} & \multirow{3}{*}{5000}\\
&&Sketch A,Sketch B,Real A,Real B,&&\\
&&Clipart A,Clipart B,Painting A,Painting B\}&&\\
\hline
\end{tabular}}
\end{center}
\end{table*}

\subsection{Implementation Details}
We provide a short description of all models compared and their implementations. All modes make use of a ResNet-101 and rescale input images to $224\times{}224$ unless stated otherwise.
\begin{description}
    \item[Glow~\cite{kingma2018glow}] is a generative flow-based model, that allows for the exact computation of the likelihood, which we use as the anomaly score at test time. We use the implementation of \footnote{\url{https://github.com/y0ast/Glow-PyTorch}}, and an architecture with three blocks of 32~layers each. Images are resized to~$32\times32$.
    \item[IC~\cite{serra2019input}] aims to correct the high likelihood that generative models tend to assign to simple inputs, such as constant color images. To this end, IC~computes the ratio between the likelihood of the generative model and a complexity score of the input image. We used the Glow described above as our generative model and the length of the PNG image encoding as the complexity estimate.
    \item[HierAD~\cite{schirrmeister2020understanding}] computes the ratio between the Glow generative model likelihood and a general background likelihood consisting of a Glow model trained on the \emph{80~Million Tiny Images} dataset~\cite{torralba200880}, provided at \footnote{\url{https://github.com/boschresearch/hierarchical\_anomaly\_detection}}. To make the method fully unsupervised, we do not use their proposed outlier loss during training.
    \item[MHRot~\cite{hendrycks2019using}] trains a multi-headed classifier to predict the correct transformation applied to an image. At test time, the classifier's softmax scores are combined for a final OOD~score. Models are trained with the default settings until convergence of the validation loss. 
    \item[DDV~\cite{marquez2019image}] aims to build an efficient latent representation by iteratively maximizing the log-likelihood of the low-dimensional latent vectors of the training images. Anomaly scores are given by the negative log-likelihood. We use our own implementation of DDV, following the settings described in its paper, i.e.,~a latent space of dimensionality 16 and a bandwidth of $10^{-2}$ \cite{marquez2019image}.
    \item[MSCL~\cite{reiss2021mean}] uses a novel contrastive loss function to fine-tune the final two blocks of a pretrained network, and combines this with an angular center loss for a final score. We used the official implementation with the learning rate set to $5\cdot 10^{-5}$, as described in the paper, and trained until convergence.
    \item[CFlow~\cite{gudovskiy2022cflow}] fits a normalizing flow network to features extracted from a pretrained network at multiple scales, conditioned on spatial information from a positional encoder. Anomaly scores are computed by aggregating the multi-scale likelihoods, upsampled to the original resolution. We again use the default hyperparameters.
    \item[DN2~\cite{bergman2020deep}] scores outliers by computing the mean distance to its 2~nearest neighbour on features extracted from the penultimate layer of a network pre-trained on ImageNet.
    \item[SSD~\cite{sehwag2021ssd}] uses contrastive learning for self-supervised representation learning. Then, it scores samples by the Mahalanobis distance
    computed at the last layer. All images were resized to $32\times32$. We use the default settings described in the official implementation.
    \item[MahaAD~\cite{rippel2021modeling}] is the Mahalanobis anomaly detector. Besides the ResNet-101, we also show results with an EfficientNet-b4 as described in~\cite{rippel2021modeling}. With the ResNets, we resize images to $224\times{}224$, while for the EfficientNet-b4 this is $380\times{}380$.
\end{description}

\subsection{Extended results}

In Table~\ref{tab:cifaruni} to Table~\ref{tab:medical} we dissect the per-task results from Table 1, reporting the AUC scores for each individual experiment and including some additional methods that were omitted from the main text for clarity.

\begin{table*}
\caption{AUC~scores for CIFAR10 experiments of \uniclass. First published~(FP) column contains the dates of first online appearance.\\
\footnotesize{* Our results}}
\label{tab:cifaruni}
\begin{center}
\resizebox{\textwidth}{!}{
 \begin{tabular}{l|cccccccccc|cc}
\hline
 & Airplane  & Automobile &  Bird &  Cat &  Deer  & Dog   &Frog &  Horse  & Ship  & Truck &  Average & FP\\
\hline
OCSVM \cite{scholkopf1999support}& 63.0 & 44.0 &64.9 &  48.7 &  73.5  & 50.0 &  72.5 &  53.3 &  64.9 &  50.8&58.5 & Dec 1999\\
AnoGAN \cite{schlegl2017unsupervised}& 67.1&54.7&52.9 &  54.5 &  65.1&   60.3 &  58.5 &  62.5  & 75.8  & 66.5&61.8 & Mar 2017\\
RCAE \cite{chalapathy2018anomaly} &72.0&63.1&71.7&   60.6  & 72.8  & 64.0 &  64.9 &  63.6  & 74.7&   74.5&68.2 & Feb 2018\\
GT \cite{golan2018deep}&74.7&95.7&78.1 &  72.4 &  87.8  & 87.8 &  83.4  & 95.5  & 93.3 &  91.3&86.0 & May 2018\\
Glow* \cite{kingma2018glow} & 76.1&	44.5&60.3&	57.3&	43.9&	55.1&	36.2&	46.4	&71.0&	46.4 & 53.7 & Jul 2018\\
LSA \cite{abati2019latent}&73.5&58.0&69.0&54.2&76.1&54.6&75.1&53.5&71.7&54.8&64.1 & Jul 2018\\
DSVDD \cite{ruff2018deep}&61.7&65.9&50.8 &  59.1  & 60.9  & 65.7 &  67.7 &  67.3  & 75.9  & 73.1&64.8 & Jul 2018\\
IIC \cite{ji2018invariant} & 68.4 & 89.4 & 49.8 & 65.3 & 60.5 & 59.1 & 49.3 & 74.8 & 81.8 & 75.7 & 67.4 & Jul 2018\\
DIM \cite{hjelm2018learning} & 72.6 & 52.3 & 60.5 & 53.9 & 66.7 & 51.0 & 62.7 & 59.2 & 52.8 & 47.6 & 57.9 & Aug 2018\\
OCGAN \cite{perera2019ocgan}&75.7&53.1&64.0  & 62.0 &  72.3  & 62.0  & 72.3   &57.5 &  82.0  & 55.4&65.6 & Mar 2019\\
MHRot \cite{hendrycks2019using} & 77.5 & 96.9 & 87.3 & 80.9 & 92.7 & 90.2 & 90.9 & 96.5 & 95.2 & 93.3 & 90.1 & Jun 2019\\
CapsNet \cite{li2019exploring} & 62.2 & 45.5&67.1&67.5&68.3&63.5&72.7&67.3&71.0&46.6&61.2 & Jul 2019\\
IC* \cite{serra2019input} &38.3&	62.0&	45.5&	61.5&	48.7&	63.9&	62.6&	63.7&	48.4&	58.8&	55.3& Jul 2019\\
E3Outlier \cite{wang2019effective} &79.4&95.3&75.4  & 73.9 &  84.1   &87.9 &  85.0  & 93.4  & 92.3  & 89.7&85.6 & Sep 2019\\
DDV* \cite{marquez2019image} & 83.2& 	58.5& 	55.4& 	56.9& 	61.2& 	57.9& 	63.3& 	57.5& 	88& 	71.2& 	65.3&  Oct 2019\\
DeepIF \cite{ouardini2019towards} & - & - & - & - & - & - & - & - & - & - & 88.2 & Oct 2019\\
CAVGA-DU \cite{venkataramanan2020attention} &65.3&78.4&76.1&74.7&77.5&55.2&81.3&74.5&80.1&74.1&73.7 & Nov 2019\\
U-Std \cite{bergmann2020uninformed} & 78.9 & 84.9 & 73.4 & 74.8 & 85.1 & 79.3 & 89.2 & 83.0 & 86.2 & 84.8 & 82.0 & Nov 2019\\
InvAE \cite{huang2019attribute} &78.5&89.8&86.1 &  77.4   &90.5  & 84.5   &89.2 &  92.9   &92.0  & 85.5&86.6 & Nov 2019\\
DROCC \cite{goyal2020drocc} &81.7&76.7&66.7  & 67.1  & 73.6 &  74.4 &  74.4   &71.4 &  80.0  & 76.2&74.2 & Feb 2020\\
DN2 \cite{bergman2020deep} & 92.8& 	97.8& 	85.3& 	85& 	94.4& 	92.7& 	93.1& 	94.4& 	95.9& 	97.3& 	92.9&  Feb 2020\\
ARAE \cite{salehi2020arae}& 72.2&43.1&69.0&55.0&75.2&54.7&70.1&51.0&72.2&40.0&60.2 & Mar 2020\\
GOAD \cite{bergman2020classification} &77.2&96.7&83.3  & 77.7  & 87.8 &  87.8  & 90.0  & 96.1 &  93.8 &  92.0&88.2 & May 2020\\
MahaAD*\textsubscript{RN101} \cite{rippel2021modeling}& 92.9&	96.4&	85.8&	85&	93.8&	91.1&	94.1&	94.8&	95.4&	96.8&	92.6 & May 2020\\
MahaAD*\textsubscript{ENB4} \cite{rippel2021modeling}& 95.1 & 97.8 & 92.3 & 91.6 & 96.5 & \textbf{96.8} & \textbf{97.6} & 96.9 & 97.4 & 98.3 & 96.0 & May 2020\\
\hline
HierAD* \cite{schirrmeister2020understanding}& 47.6&	63.4	&63.2&	59.0&	79.2&	64.3&	77.5&	66.4&	61.6&	59.8&	64.2 & Jun 2020\\
CSI \cite{tack2020csi}&89.9 &\textbf{99.9}&93.1&86.4&93.9&93.2  & 95.1&98.7  & 97.9&95.5&94.3 & Jul 2020\\
Puzzle-AE \cite{salehi2020puzzle} & 78.9 & 78.1 & 70.0 & 54.9 & 75.5 & 66.0 & 74.8 & 73.3 & 83.3 & 70.0 & 72.5 & Aug 2020\\
PANDA \cite{reiss2021panda} & \textbf{97.4} & 98.4 & 93.9 & 90.6 & \textbf{97.5} & 94.4 & 97.5 & 97.5 & 97.6 & 97.4 &96.2 & Oct 2020\\
ConDA \cite{sohn2020learning} & 90.9 & 98.9 & 88.1 & 83.1 & 89.9 & 90.3 & 93.5 & 98.2 & 96.5 & 95.2 & 92.5 & Nov 2020\\
MKD \cite{salehi2021multiresolution} & 90.5 & 90.4 & 79.7 & 77.0 & 86.7 & 91.4 & 89.0 & 86.8 & 91.5 & 88.9 & 87.2 & Nov 2020\\
SSD \cite{sehwag2021ssd}&82.7&98.5&84.2 &  84.5  & 84.8&   90.9&   91.7 & 95.2 &  92.9&   94.4&90.0 & Mar 2021\\
SSL \cite{xiao2021we} &94.8&96.4&88.3&87.6&92.7&94.2 &  96.4&94.3 & 96.1&97.0&93.8 & May 2021\\
MTL \cite{mohseni2021multi} & 84.3 & 96.0 & 87.7 & 82.3 & 91.0 & 91.5 & 91.1 & 96.3 & 96.3 & 92.3 & 90.9 & Jun 2021\\
MSCL* \cite{reiss2021mean} & 97& 	98.6& 	94.6& 	92.2& 	97.1& 	96.4& 	96.5& 	97.9& 	98.4& 	\textbf{98.6}& 	\textbf{96.7}&  Jun 2021\\
OODformer \cite{koner2021oodformer} & 92.3 &99.4& \textbf{95.6}& \textbf{93.1} &94.1 &92.9 &96.2 &\textbf{99.1}& \textbf{98.6}& 95.8 & 95.7 & Jul 2021\\
DaA \cite{hou2021divide} & - & - & - & - & - & - & - & - & - & - & 75.3 & Jul 2021\\
CFlow* \cite{gudovskiy2022cflow} & 69.4& 	83.5& 	68& 	73.9& 	84.7& 	77.9& 	84.4& 	78.6& 	80.3& 	84.2& 	78.5& Jul 2021\\
\hline
\end{tabular}}
\end{center}
\end{table*}

\begin{table*}
\caption{AUC scores for CIFAR100 experiments of\uniclass. \\
\footnotesize{* Our results}}
\label{tab:cifar100}
\begin{center}
\resizebox{\textwidth}{!}{
\begin{tabular}{l|cccccccccccccccccccc|c}
\hline
 & 0 & 1 & 2 & 3 & 4 & 5 & 6 & 7 & 8 & 9 & 10 & 11 & 12 & 13 & 14 & 15 & 16 & 17 & 18 & 19 & Mean\\
\hline
Glow* \cite{kingma2018glow} & 60.7	&59.4&	25.4&	65.7&	45.5&	66.9&	66.1&	46.0	&46.0&	64.8&75.5&	51.1&	54.0&	48.8&	50.6&	50.2&	52.8&	50.1&	44.1&	53.3&	53.8 \\
IC* \cite{serra2019input} & 61.2&	53.9&	44.4&	44.4&	48.3&	46.4&	41.9&	51.2&	72.0&	58.0&	48.7&	68.3&	69.8&	51.6&	56.1&	62.0&	62.4&	68.8&	59.5&	48.8&	55.9 \\
OC-SVM \cite{scholkopf1999support}&	68.4 &	63.6 &	52 &	64.7 &	58.2 &	54.9 &	57.2 &	62.9 &	65.6 &	74.1 &	84.1 &	58 &	68.5 &	64.6 &	51.2 &	62.8 &	66.6 &	73.7 &	52.8 &	58.4 &	63.1 \\
DAGMM \cite{zong2018deep}&	43.4 &	49.5 &	66.1 &	52.6 &	56.9 &	52.4 &	55 &	52.8 &	53.2 &	42.5 &	52.7 &	46.4 &	42.7 &	45.4 &	57.2 &	48.8 &	54.4 &	36.4 &	52.4 &	50.3 &	50.6 \\
DSEBM \cite{zhai2016deep}&	64 &	47.9 &	53.7 &	48.4 &	59.7 &	46.6 &	51.7 &	54.8 &	66.7 &	71.2 &	78.3 &	62.7 &	66.8 &	52.6 &	44 &	56.8 &	63.1 &	73 &	57.7 &	55.5 &	58.8 \\
DDV* \cite{marquez2019image} & 58.3 & 	58 & 	70.6 & 	75.3 & 	72.2 & 	60.3 & 	65.4 & 	61.4 & 	63.8 & 	72 & 	77 & 	55.5 & 	82.8 & 	53.4 & 	61.4 & 	58.6 & 	51.9 & 	87.5 & 	64.5 & 	72.3 & 	66.1  \\
HierAD* \cite{schirrmeister2020understanding} & 68.7&	59.5&	76.5&	35.9&	59.7&	31.6&	48.5&	59.6&	78.4&	65.1&	76.9&	67.6&	77.1&	55.1&	59.1&	63.2&	69.6&	80.1&	58.4&	57.7&	62.4\\
DVSDD \cite{ruff2018deep} & 66& 	60.1& 	59.2& 	58.7& 	60.9& 	54.2& 	63.7& 	66.1& 	74.8& 	78.3& 	80.4& 	68.3& 	75.6& 	61& 	64.3& 	66.3& 	72& 	75.9& 	67.4& 	65.8& 	67.0\\
GOAD \cite{bergman2020classification}&	73.9 &	69.2 &	67.6 &	71.8 &	72.7 &	67 &	80 &	59.1 &	79.5 &	83.7 &	84 &	68.7 &	75.1 &	56.6 &	83.8 &	66.9 &	67.5 &	91.6 &	88 &	82.6 &	74.5 \\
MHRot \cite{hendrycks2019using} & 77.6& 	72.8& 	71.9& 	81& 	81.1& 	66.7& 	87.9& 	69.4& 	86.8& 	91.7& 	87.3& 	85.4& 	85.1& 	60.3& 	92.7& 	70.4& 	78.3& 	93.5& 	89.6& 	88.1& 	80.1 \\
SSD* \cite{sehwag2021ssd} & 76.5&	79.6&	88.7&	73.4&	91.1&	72.4&	73.9&	79.8&	80.7&	86.0&	72.3	&79.4	&83.1&	74.5&	87.3&	74.4&	79.9&	90.9&	83.3&	80.7&	80.4\\
ConDA \cite{sohn2020learning} & 82.9& 	84.3& 	88.6& 	86.4& 	92.6& 	84.5& 	73.4& 	84.2& 	87.7& 	94.1& 	85.2& 	87.8& 	82& 	82.7& 	93.4& 	75.8& 	80.3& 	97.5& 	94.4& 	92.4& 	86.5\\
CSI \cite{tack2020csi} & 86.3& 	84.8& 	88.9& 	85.7& 	93.7& 	81.9& 	91.8& 	83.9& 	91.6& 	95& 	94& 	90.1& 	90.3& 	81.5& 	94.4& 	85.6& 	83& 	97.5& 	95.9& 	95.2& 	89.6\\
MKD* \cite{salehi2021multiresolution} & 90.3&	89.7&	90.1&	89.9&	89.8&	90.2&	89.7&	90.3&	90.0&	89.5&	88.5&	90.2&	91.0&	89.6&	89.0&	89.8&	90.4&	88.9&	90.1&	90.7&	89.9\\
DN2* \cite{bergman2020deep} & 88.3 & 	85.6 & 	95.1 & 	95.1 & 	94.4 & 	93.8 & 	94.4 & 	87.3 & 	92.7 & 	91.4 & 	95.8 & 	87.4 & 	88.1 & 	79.3 & 	95.8 & 	78.6 & 	84.1 & 	96.6 & 	91.1 & 	90.4 & 	90.3  \\
PANDA \cite{reiss2021panda} & 91.5& 	92.6& \textbf{	98.3}& 	96.6& 	96.3& 	94.1& 	96.4& 	91.2& 	94.7& 	94& 	96.4& 	92.6& 	93.1& 	89.4& 	98& 	89.7& 	92.1& 	97.7& 	94.7& 	92.7& 	94.1\\
MSCL* \cite{reiss2021mean}& \textbf{95.8} & 	\textbf{95.2} & 	97.6 & 	\textbf{98.3} & 	97.1 & 	\textbf{96.9} & \textbf{98.3} & 	\textbf{94.7} & 	\textbf{97.6} & 	\textbf{97.9} & 	\textbf{97.4} & 	\textbf{96.3} & 	\textbf{94.9} & 	\textbf{91.7} & \textbf{98.3} & 	\textbf{92.7} & 	\textbf{93.1} & \textbf{98.3} & 	\textbf{97.9} & 	\textbf{97.4} & 	\textbf{96.4} \\
CFlow* \cite{gudovskiy2022cflow} & 75.3 & 	67.2 & 	76 & 	76 & 	76.6 & 	71.7 & 	76.5 & 	57.9 & 	79.8 & 	83.7 & 	91.5 & 	70.4 & 	74.3 & 	63.1 & 	71.5 & 	64.8 & 	70.3 & 	90.6 & 	64.9 & 	62 & 	73.2 \\
\hline
MahaAD*\textsubscript{RN101} \cite{rippel2021modeling}& 91.9&	89.5&	96&	95.3&	94.7&	91.1&	95.2&	89.5&	93.6&	93.7&	95.4&	90.6&	91.4&	84.3&	96.7&	84.5&	87.7	&97.1&	94.4&	92.8&	92.3 \\
MahaAD*\textsubscript{ENB4} \cite{rippel2021modeling}& 93.2 & 92.8 & 96.7 & 97.8 & \textbf{97.2} & 95.4 & 98.0 & 92.6 & 95.9 & 94.9 & 95.8 & 93.0 & 93.0 & 89.2 & 97.8 & 89.1 & 91.7 & 97.5 & 96.2 & 94.8 & 94.6 \\
\hline
\end{tabular}
}
\end{center}
\end{table*}

\begin{table*}
\caption{AUC scores for \shiftlowres.\\
\footnotesize{* Our results}}
\label{tab:smallood}
\begin{center}
\makebox[\textwidth][c]{
\begin{tabular}{l|c}
\hline
 & CIFAR10:SVHN \\
\hline
CFlow* \cite{gudovskiy2022cflow} & 6.6 \\
Glow \cite{schirrmeister2020understanding} & 8.8 \\
DSVDD \cite{ruff2018deep} & 14.5\\
MKD* \cite{salehi2021multiresolution} & 26.8\\
DDV* \cite{marquez2019image} & 47.9 \\
EBM \cite{du2019implicit} & 63.0  \\
DN2* \cite{bergman2020deep} & 57.4 \\
VAEBM \cite{xiao2020vaebm} & 83.0  \\
MSCL* \cite{reiss2021mean} & 88.3  \\
TT \cite{nalisnick2019detecting} & 87.0\\
LLRe \cite{xiao2020likelihood} & 87.5 \\
BIVA \cite{havtorn2021hierarchical} & 89.1   \\
NAE \cite{yoon2021autoencoding} & 92.0 \\
HierAD \cite{schirrmeister2020understanding} & 93.9\\
IC \cite{serra2019input} & 95.0\\
GOAD \cite{bergman2020classification} & 96.3\\
SVD-RND \cite{choi2019novelty}&96.4 \\
MHRot \cite{hendrycks2019using} & 97.8\\
DoSE \cite{morningstar2021density} & 97.3\\
CSI \cite{tack2020csi} & 99.8 \\
SSD \cite{sehwag2021ssd} & 99.6 \\
MTL \cite{mohseni2021multi} & 99.9 \\
WAIC \cite{morningstar2021density} & 14.3\\
WAIC \cite{choi2018waic} & \textbf{100} \\
\hline
MahaAD*\textsubscript{RN101} \cite{rippel2021modeling}& 94.3 \\
MahaAD*\textsubscript{ENB4} \cite{rippel2021modeling}& 96.2 \\
\hline
\end{tabular}}
\end{center}
\end{table*}

\begin{table*}
\caption{AUC scores for \shifthighres{} using \texttt{Real-A} as the in-distribution. QD:~quickdraw, IG:~infograph, SK:~sketch, CA:~Clipart, PN:~Painting. A~is the set without semantic shift, and B~with semantic shift.\\
\footnotesize{* Our results}}
\label{tab:domainReal}
\begin{center}
\makebox[\textwidth][c]{
\begin{tabular}{l|cccccccccc|c}
\hline
 & QDa & QDb & IGa & IGb & SKa & SKb & CAa & CAb & PNa & PNb & Mean\\
\hline
MSCL* \cite{reiss2021mean} & 33.8 & 	32.9 & 	68.6 & 	67.1 & 	54.7 & 	58.9 & 		58.3 & 	61.3 & 	72.4 & 	75.5 &  	58.3  \\
SSD* \cite{sehwag2021ssd} & 40.3&	40.4&	69.0&	69.6&	68.9&	73.9&	53.1&	58.8&	77.6&	83.3&	64.0\\
MKD* \cite{salehi2021multiresolution} & 24.2&	23.1&	56.6&	52.7&	47.2&	47.3 &	49.4&	47.3&	68.6&	70.4 & 48.9\\
DDV* \cite{marquez2019image} & 87.9 & 	90.9 & 	56 & 	54.6 & 	62.6 & 	64.6 & 		62.1 & 	64.8 & 	52 & 	59.4 & 	64.0  \\
DN2* \cite{bergman2020deep} &50.4 & 	50.8 & 	76.2 & 	74 & 	69.1 & 	73.7 & 		70.7 & 	\textbf{74.7}& 	\textbf{79.7} & 	\textbf{85.0} & 70.4 \\
MHRot* \cite{hendrycks2019using} & 71.6&	71.6&	48.7&	50.1&	63.8	&64.4& 60.2&	61.5&	55.4&	57.0 &59.7\\
Glow* \cite{kingma2018glow} & 3.2&	3.0	&54.8	&51.0&	19.5&	20.9&	37.1&	33.4&	66.6&	67.0&	36.9 \\
IC* \cite{serra2019input} &89.9&	90.4&	66.4&	68.8&	69.5&	68.8	&64.4&	66.3&	55.9&	55.7&	68.0 \\
HierAD* \cite{schirrmeister2020understanding} &\textbf{95.5}	&\textbf{95.7}&	36.6&	40.6&	\textbf{84.9}&	\textbf{82.7}&		51.5	&58.3&	41.6&	41.6&	61.8\\
CFlow* \cite{gudovskiy2022cflow} & 46.6 & 	47 & 	52.2 & 	49.6 & 	48.9 & 	51.1 & 		62.3 & 	62.9 & 	58 & 	57.6 & 53.6 \\
\hline
MahaAD*\textsubscript{RN101} \cite{rippel2021modeling}& 72.9&	71.3&	\textbf{81.6} &	\textbf{80.8}&	64.2&	65.5&	70.3&	70&	66	&69.2	&71.2\\
MahaAD*\textsubscript{ENB4} \cite{rippel2021modeling}& 79.7&	80.4&	76.3&	76.9&	73.8&	76.3 & \textbf{71.0} & 73.5 & 70.5 & 77.5 & \textbf{75.6}\\
\hline
\end{tabular}}
\end{center}
\end{table*}

\begin{table*}
\caption{AUC scores for \shifthighres{} using \texttt{Infograph-A} as the in-distribution.\\
\footnotesize{* Our results}}
\label{tab:domainIG}
\begin{center}
\makebox[\textwidth][c]{
\begin{tabular}{l|cccccccccc|c}
\hline
 & QDa & QDb & SKa & SKb & REa & REb & CAa & CAb & PNa & PNb & Mean\\
\hline
MSCL* \cite{reiss2021mean} & 91.9 & 	91.9 & 		83.9 & 	84.3 & 	92.7 & 	92.8 & 	87.3 & 	86.5 & \textbf{	96.3} & 	\textbf{96.2} & 90.4\\
SSD* \cite{sehwag2021ssd} & 35.1&	33.5	&	67.9&	69.1&	56.7	&57.7&	69.4&	69.3&	57.3&	58.5&	57.3\\
MKD* \cite{salehi2021multiresolution} & 83.0	&82.4&	81.7&	80.4&	88.9&	91.0&	84.5&	82.5&	95.6&	95.2&	83.0 \\
DDV* \cite{marquez2019image} & 59.5 & 	72.3 & 		56.3 & 	63.4 & 	69.7 & 	75.4 & 	46.6 & 	54.3 & 	70.3 & 	69.9 & 	63.8 \\
DN2* \cite{bergman2020deep} & 75.1 & 	75.7 & 		75.1 & 	76.8 & 	82.7 & 	88.1 & 	80.1 & 	79.5 & 	91.2 & 	92.1 & 	81.6\\
MHRot* \cite{hendrycks2019using} & 94.9&	95.2 &	88.5&	88.7&	87.6&	87.9&	\textbf{89.3}&	\textbf{89.7}&	88.6&	89.4&	86.7 \\
Glow* \cite{kingma2018glow} &0.7&	0.6	&12.3&	14.0&	50.7&	49.9&	35.3&	30.6&	69.2&	69.5	&34.4\\
IC* \cite{serra2019input} &94.1&	94.4	&64.8&	63.5&	42.9&	44.8&	60.3&	62.4&	46.7	&46.8	&61.3\\
HierAD* \cite{schirrmeister2020understanding}&\textbf{99.8}&\textbf{99.8}&	\textbf{93.8}&	\textbf{92.7}&	83.1&	83.3&	80.8&	83.1&	77.6&	77.6&	84.1 \\
CFlow* \cite{gudovskiy2022cflow} & 68.8 & 	69 & 		64.9 & 	65.2 & 	74.7 & 	74.9 & 	75.7 & 	75.9 & 	74.5 & 	73.6 & 	71.7 \\
\hline
MahaAD*\textsubscript{RN101} \cite{rippel2021modeling}& 92.3&	92.1&	78.1&	77.6&	88.1&	88.4&	81.5&	80.3&	90.9	&91.2 & 86.1\\
MahaAD*\textsubscript{ENB4} \cite{rippel2021modeling}& 94.5 & 94.8 & 89.5& 89.0 & \textbf{93.6} & \textbf{94.7} & 87.4 & 87.1 & 94.9 & 95.4 & \textbf{92.1}\\
\hline
\end{tabular}}
\end{center}
\end{table*}

\begin{table*}
\caption{AUC~scores for \uniano. HN is hazelnut, MN is metal nut, TB is toothbrush and TS is transistor. \\
\footnotesize{* Our results}}
\label{tab:mvtec}
\begin{center}
\resizebox{\textwidth}{!}{
\begin{tabular}{l|ccccccccccccccc|c}
\hline
 & Carpet& Grid& Leather &Tile &Wood &Bottle& Cable& Capsule& HN &MN &Pill &Screw& TB& TS& Zipper &Mean\\
\hline
AVID \cite{sabokrou2018avid} &	70 &	59 &	58 &	66 &	83 &	88 &	64 &	85 &	86 &	63 &	86 &	66 &	73 &	58 &	84 &	73 \\
AESSIM \cite{bergmann2018improving} &	67 &	69 &	46 &	52 &	83 &	88 &	61 &	61 &	54 &	54 &	60 &	51 &	74 &	52 &	80 &	63 \\
AEL2 \cite{bergmann2018improving} &	50 &	78 &	44 &	77 &	74 &	80 &	56 &	62 &	88 &	73 &	62 &	69 &	98 &	71 &	80 &	71 \\
AnoGAN \cite{schlegl2017unsupervised} &	49 &	51 &	52 &	51 &	68 &	69 &	53 &	58 &	50 &	50 &	62 &	35 &	57 &	67 &	59 &	55 \\
LSA \cite{abati2019latent} &	74 &	54 &	70 &	70 &	75 &	86 &	61 &	71 &	80 &	67 &	85 &	75 &	89 &	50 &	88 &	73 \\
CAVGA-DU \cite{venkataramanan2020attention} &	73 &	75 &	71 &	70 &	85 &	89 &	63 &	83 &	84 &	67 &	88 &	77 &	91 &	73 &	87 &	78 \\
DSVDD \cite{ruff2018deep} &	54 &	59 &	73 &	81 &	87 &	86 &	71 &	69 &	71 &	75 &	77 &	64 &	70 &	65 &	74 &	72 \\
VAE-grad \cite{dehaene2020iterative} &	67 &	83 &	71 &	81 &	89 &	86 &	56 &	86 &	74 &	78 &	80 &	71 &	89 &	70 &	67 &	77 \\
GT \cite{golan2018deep} &	46 &	61.9 &	82.5 &	53.9 &	48.2 &	74.3 &	84.8 &	67.8 &	33.3 &	82.4 &	65.2 &	44.6 &	94 &	79.8 &	87.4 &	67.1 \\
Puzzle-AE \cite{salehi2020puzzle} &	65.7 &	75.4 &	72.9 &	65.5 &	89.5 &	94.2 &	87.9 &	66.9 &	91.2 &	66.3 &	71.6 &	57.8 &	97.8 &	86 &	75.7 &	77.6 \\
MKD \cite{salehi2021multiresolution} &	79.3 &	78 &	95.1 &	91.6 &	94.3 &	99.4 &	89.2 &	80.5 &	98.4 &	73.6 &	82.7 &	83.3 &	92.2 &	85.6 &	93.2 &	87.7 \\
MSCL* \cite{reiss2021mean} &92.6& 	53.8& 	98& 	97.2& 	91.2& 	98.7& 	88.8& 	87.4& 	94.1& 	85& 	68.8& 	63.7& 	87.5& 	93.2& 	96.4& 	86.4 \\
SSD* \cite{sehwag2021ssd} &	53.4 &	33.5 &	61.4 &	61.9 &	44.9 &	78.3 &	62.7 &	60.2 &	62.2 &	69.4 &	76.6 &	59.5 &	99.8 &	88.5 &	74.8 &	65.8 \\
DDV* \cite{marquez2019image} &	80.3& 	42& 	55.1& 	47.4& 	46.4& 	99.7& 	66.1& 	77.2& 	64.2& 	81& 	71.9& 	53.6& 	64.1& 	77.8& 	56& 	65.5 \\
DN2* \cite{bergman2020deep} &	90.3& 	56.4& 	98.9& 	99.2& 	96.8& 	99.2& 	82& 	84.4& 	92.9& 	83.6& 	69.5& 	66.4& 	88.1& 	91.3& 	93.8& 	86.2 \\
MHRot* \cite{hendrycks2019using} &	47.8 &	58.9 &	75 &	51.2 &	90.2 &	82 &	79.9 &	59 &	73.6 &	75.7 &	64.9 &	36.6 &	86.9 &	86.5 &	93.4 &	70.8 \\
Glow* \cite{kingma2018glow} &	72.9 &	\textbf{98.3} &	94.1 &	83.7 &	96.9 &	96.6 &	83.3 &	67.1 &	90.5 &	62.4 &	84.8 &	31.8 &	87.6 &	88.4 &	91.3 &	82.0 \\
IC* \cite{serra2019input} &	69.7 &	75.6 &	94.3 &	71.2 &	78.1 &	96.0 &	85.8 &	63.3 &	64.9 &	77.0 &	67.9 &	29.7 &	85.8 &	89.5 &	54.9 &	73.6 \\
HierAD* \cite{schirrmeister2020understanding} &	73.4 &	95.3 &	95.5 &	84.5 &	97.5 &	97.3 &	86.5 &	70.0 &	75.0 &	73.6 &	74.2 &	26.2 &	98.6 &	92.5 &	84.1 &	81.6 \\
SPADE \cite{cohen2020sub}&	- &	- &	- &	- &	- &	- &	- &	- &	- &	- &	- &	- &	- &	- &	- &	85.5 \\
FAVAE \cite{dehaene2020iterative}  &	67.1 &	97 &	67.5 &	80.5 &	94.8 &	\textbf{99.9} &	95 &	80.4 &	99.3 &	85.2 &	82.1 &	83.7 &	95.8 &	93.2 &	97.2 &	87.9 \\
AEsc \cite{collin2021improved} &	89 &	97 &	89 &	99 &	95 &	98 &	89 &	74 &	94 &	73 &	84 &	74 &	\textbf{100} &	91 &	94 &	89 \\
DaA \cite{hou2021divide} &	86.6 &	95.7 &	86.2 &	88.2 &	98.2 &	97.6 &	84.4 &	76.7 &	92.1 &	75.8 &	90 &	\textbf{98.7} &	99.2 &	87.6 &	85.9 &	89.5 \\
CFlow* \cite{gudovskiy2022cflow} & \textbf{99.3}& 	93.3& 	\textbf{100}& 	99.2& \textbf{98.4}& 	\textbf{99.9}& 	\textbf{95.5}& 	90.9& 	\textbf{99.7}& 	\textbf{99.5}& \textbf{	92.3}& 	83& 	92.2& 	93.9& 	\textbf{98.7}& 	\textbf{95.7} \\
\hline
MahaAD*\textsubscript{RN101} \cite{rippel2021modeling} &	79.5 &	59.6 &	99.3 &	\textbf{100} &	98.2 &	99.3 &	91.6 &	93.8 &	99.4 &	93.4 &	90.6 &	72.1 &	98.6 &	96.1 & 97.9 &	91.3 \\
MahaAD*\textsubscript{ENB4} \cite{rippel2021modeling} &	98.6 &	78.8 &	99.7 &	\textbf{100} &	96.1 &	99.8 &	93.5 &	\textbf{97.0} &	99.0 &	93.9 &	90.3 &	78.6 &	96.7 &	\textbf{96.5} &	97.7 &	94.4 \\
\hline
\end{tabular}
}
\end{center}
\end{table*}

\begin{table*}
\caption{AUC scores for \unimed.\\
\footnotesize{* Our results}}
\label{tab:medical}
\begin{center}
\scalebox{1}{
\begin{tabular}{l|ccccccc}
\hline
 & OCT & Chest & NIH & DRD1 & DRD2 & DRD3 & DRD4\\
\hline
IF \cite{liu2008isolation} &&&&&&&44.0\\
AnoGAN \cite{schlegl2017unsupervised}&&&&&&&44.2\\
DSEBM \cite{zhai2016deep} &&&&&&&43.1\\
DAGMM \cite{zong2018deep} &&&&&&&52.0\\
Glow \cite{kingma2018glow} & 44.8 & 54.6 &  &  & & & \\
GT \cite{bergman2020classification} &  &  & 79.2 &&&&\\
DSVDD \cite{ruff2018deep} & 77.4 & 66.6 & 81.8 &&&&46.4\\
DeepIF \cite{ouardini2019towards} &&&&&&&74.5\\
DDV \cite{marquez2019image} & 86.7 & 	79.9 & 	57.7 & 	45.3 & 	48.9 & 	50.2 & 	53.4\\
GAOCC \cite{tang2019abnormal} &  &  & 83.4 &&& \\
MemDAE \cite{bozorgtabar2020salad}&  &  & 87.8 \\
MSCL* \cite{reiss2021mean} & 94.1 & 	93.3 & 	81.9 & 	52 & 	\textbf{55.8} & 	68.2 & 	81.1 \\
SSD* \cite{sehwag2021ssd} & 59.4&	94.5&	74.2&	47.5&	50.6&	54.8&	71.4\\
MKD* \cite{salehi2021multiresolution}& 94.9&	95.8&	\textbf{88.0}&	53.7	&54.6&	60.7&	75.5\\
DN2* \cite{bergman2020deep} & 94.1 & 	96.9 & 	81.2 & 	\textbf{54.4} & 	55.6 & 	\textbf{69.4} & 	\textbf{85.4} \\
MHRot* \cite{hendrycks2019using} & 87.7&	96.2&	81.8&	49.0&	50.2&	52.7&	65.3\\
Glow* \cite{kingma2018glow}& 62.3&	49.8	&65.0&	52.2&	47.5&	54.7&	59.5\\
IC* \cite{serra2019input}& 83.4	&91.6&	56.7	&47.5	&52.1&	58.2&	66.2\\
HierAD* \cite{schirrmeister2020understanding} &94.3	&99.0&	79.8	&52.1&	51.7&	57.5&73.5\\
CFlow* \cite{gudovskiy2022cflow} & 76.4 & 	81.7 & 	78.7 & 	53.2 & 	55.1 & 	61.3 & 	75.1 \\
\hline
MahaAD*\textsubscript{RN101} \cite{rippel2021modeling}& 98&	\textbf{99.8}&	84.6&	52.1&	52	&63.6&	79.9 \\
MahaAD*\textsubscript{ENB4} \cite{rippel2021modeling}& \textbf{98.7} & \textbf{99.8}&84.2 & 49.9 & 55.0 & 66.3 & 81.3\\
\hline
\end{tabular}}
\end{center}
\end{table*}

\end{document}